\documentclass[3p, 12pt]{elsarticle}

\usepackage{lineno,hyperref}

\modulolinenumbers[5]

\journal{Journal of \LaTeX\ Templates}

\biboptions{authoryear}

\usepackage[utf8x]{inputenc}
\usepackage[T1]{fontenc}
\usepackage{comment}
\usepackage[colorinlistoftodos]{todonotes}
\usepackage{xcolor}
\usepackage{graphicx}
\usepackage{amsmath,amsthm}
\usepackage{amssymb}
\usepackage{adjustbox}
\usepackage{algorithm}
\usepackage{algorithmic, eqparbox, array}
\usepackage{soul}
\usepackage{multirow}
\usepackage{subcaption}

\definecolor{darkgreen}{RGB}{0, 150, 0}

\newcommand{\mycomment}[1]{}
\definecolor{orange}{rgb}{1,0.5,0}

\newcommand{\inserted}[1]{\textcolor{black}{{#1}}}

\newcommand{\moved}[1]{\textcolor{black}{{#1}}}

\newcommand{\modified}[1]{\textcolor{black}{{#1}}}
\newcommand{\insertedd}[1]{\textcolor{black}{{#1}}}

\newcommand{\AggregatedGraph}{G^+}
\newcommand{\AggregatedGraphVertex}[1][]{V^+}
\newcommand{\AggregatedGraphEdge}[1][]{E^+}
\newcommand{\ThresholdGraph}[1][]{G^{#1}_\theta}
\newcommand{\ThresholdEdges}[1][]{E^{#1}_\theta}
\newcommand{\IntersectedGraph}{G^{\cap}}

\newcommand{\ShortNameMiningAlgorithm}{MMDG}
\newcommand{\FullNameThresholdingAlgorithm}{Refinement}

\theoremstyle{definition}
\newtheorem{definition}{Definition}
\theoremstyle{plain}

\newenvironment{rmk}{\noindent{\bf Remark. }} 

\begin{document}

\begin{frontmatter}

\title{
{\modified{Learning Alternative Ways of Performing a Task}}
}

\author{D. Nieves, MJ. Ramírez-Quintana, C. Monserrat\footnote{Corresponding author: cmonserr@dsic.upv.es}, C. Ferri, J. Hernández-Orallo}
\address{Valencian Research Institute for Artificial Intelligence (VRAIN), \\ Universitat Politècnica de València, Spain}




\begin{abstract}

A common way of learning to perform a task is to observe how it is carried out by experts. However, it is well known that for most tasks there is no unique way to perform them.  This is especially 	noticeable the more complex the task is because  factors such as the skill or the know-how  of the expert may well affect the way she solves the task. In addition,  learning from experts also suffers of having a small set of training examples generally coming from several experts (since experts are usually a limited and expensive resource), being all of them positive examples (i.e. examples that represent successful executions of the task). Traditional machine learning techniques are not useful in such scenarios, \modified{as they require  extensive training data.} 
\modified{Starting from very few executions of the task presented as 
activity sequences, we introduce} a novel inductive approach for learning  multiple models, \modified{with each one representing an alternative strategy} of performing a task. By an iterative process based on generalisation and  specialisation,  
we learn the underlying patterns that capture the different styles of performing a task exhibited by the examples. 
We illustrate our approach on \modified{two common activity recognition tasks: a surgical skills training task and a cooking domain}. We evaluate the inferred models with respect to two metrics that measure  how well the models represent the examples and capture the different forms of executing a task showed by the examples. \inserted{We compare our results with the traditional process mining approach and} show that a small set of meaningful examples is enough to obtain patterns that \modified{capture the different strategies that are followed to solve the tasks}. 

\end{abstract}


\end{frontmatter}

\section{Introduction} 
Nowadays, 
\modified{humans learn the execution of a complex task through} three steps: \modified{studying} the description of the task, 
watching video executions of the task (usually performed by experts) or real-life demonstrations of it, and, finally, executing  the task under expert supervision \moved{several times} \citep{ericsson2009development}. This way of acquiring the skills needed to perform a task is expensive 
in time and resources. Besides, the lack of continuous supervision may induce mistakes because of 
the limited experience of the 
\modified{operators} 
or \modified{lack of}
attention because of the repetitiveness of the task. In this sense, artificial intelligence (AI), and \modified{machine learning (ML) in particular}, is making it possible to help people in their daily lives by learning models about their tasks. These models can then be integrated into direct assistance systems, such as \modified{task training}  learning environments, supervisory contexts in order to avoid human mistakes, or machine-human collaboration contexts, where the machine is used as an assistant for a complex task. Research in this direction can be found in \inserted{fields such as} {\em Ambient Intelligence} \citep{camacho2014ontology}, {\em Context-aware systems} \citep{hong2009context}, {\em Advanced Driving Assistants} \citep{vskrjanc2018evolving} or {\em Surveillance systems} \citep{kardas2017svas}, to name a few.

However, traditional ML techniques require large volumes of data in order to infer  models of common tasks.  
\inserted{In many areas, such as surgery,}
the access to training examples  is very costly, as these are very complex tasks that very few people know how to perform \inserted{or require specific permissions or instrumental}. In these cases, the 
\modified{knowledge acquisition has to be done from a small set of examples, with examples being a series of  steps that an expert} carried out to complete the task from  beginning to end. Besides, this learning process must consider that any process can be performed in several ways; experts can present different styles when executing \modified{the task} (in many cases, involuntarily) and can contain noise (interpreted as \modified{non-essential} activities of the task). 

Other approaches \citep{yang2017medical,blum2008workflow} can generalise a pattern of the task from a few positive examples. However, they only can learn one general pattern from the set of provided examples. The same case occurs with methods such as \modified{`Fuzzy Mining' \citep{gunther2007fuzzy} or the Alpha-algorithm} \citep{alves2004process}. The previous solutions can lead to incorrect learning of the task and even dangerous learning. Let us illustrate the case with an example. Imagine a scenario where a system has to supervise the task of cooking a carrot soup. Table \ref{tbl:cooking_vocabulary} shows the vocabulary of activities involved in this task. We have two examples of how to cook this dish (expressed as sequences of activities): ``{\sf SABCDFGHIMOPT}''
and ``{\sf SABCEFGJKLNOPT}''. These examples show two different 
forms of cooking: using a microwave or a kitchen stove. 
If we want to learn a general rule from these two examples, a possible generalisation is the expression ``{\sf SABC \{D|E\}FG\{HIM|JKLN\}OPT}'', where ``|'' means a disjunction between two activity groups for the portion of the task delimited by curly brackets. The two given examples are covered by the model, but there are other valid sequences of activities according to this model that are not safe. For instance, the sequence ``{\sf SABCEFGHIMOPT}'', where the person takes a metal pot,  put the ingredients in, and, then, put the metal pot into the microwave to prepare the soup. We need a learning procedure and a representation language that is able to find the right trade-off between an overgeneralisation covering almost everything and an overspecialisation that is simply  the sequence composed by the disjunction of the two examples ``{\sf \{SABCDFGHIMOPT|SABCEFGJKLNOPT\}}''. Additionally, there can be examples that contain noise, for instance the sequence ``{\sf SABBCDFGHIMOPT}'' \inserted{is wrong, as} the person washes the carrot twice (activity {\sf B}) before \modified{cutting} it. To avoid an overgeneralisation, given that we do not have negative examples, the model should be as close as possible to the observed examples but, at the same time, avoiding the overspecialisation and the noise as, for instance, the pattern ``{\sf SABC\{DFGHIM|EFGJKLN\}OPT}''.

\begin{table}[!t]
\centering
\resizebox{10cm}{!}{%
\begin{tabular}{cl}
\multicolumn{1}{l}{Activity Code} & Activity Description \\ \hline
{\sf S} & Start the task \\
{\sf A} & Peel the carrot \\
{\sf B} & Wash the carrot \\
{\sf C} & Cut the carrot into pieces \\
{\sf D} & Take a glass pot \\
{\sf E} & Take a metal pot \\
{\sf F} & Fill the pot with water \\
{\sf G} & Add salt \\
{\sf H} & Put the pot into the microwave. \\
{\sf I} & Turn on the microwave at the maximum level (15 minutes). \\
{\sf J} & Put the pot on the stove \\
{\sf K} & Turn on the stove at the maximum level (10 minutes). \\
{\sf L} & Reduce the stove at the minimum level (30 minutes). \\
{\sf M} & Turn off the microwave. \\
{\sf N} & Turn off the stove. \\
{\sf O} & Take a deep bowl. \\
{\sf P} & Add two tablespoons of soup in the dish. \\
{\sf T} & Finish the task
\end{tabular}%
}
\caption{Vocabulary of activities for cooking a carrot soup. }
\label{tbl:cooking_vocabulary}
\end{table}


In this paper, we propose a new 
inductive method whose  advantages
are  summarised as follows:
\begin{itemize}
    \item 
    The method is able to identify and extract different models of a task from  a reduced set of executions performed by experts. These executions are presented as sequences of activities. 
    
    \item Our approach relies on a  representation language based on graphs for characterising both the examples and the models. 
    Examples and  models are represented as dependency graphs, 
    a kind of representation that simplifies an activity sequence, by only considering the consecutive dependencies directly, and representing each activity just once in the graph. Dependency graphs are hence a more concise {\em non-univocal} simplification of an activity sequence.  Similar formalisms based on graphs can be found in the workflow learning literature \citep{yang2017medical,gunther2007fuzzy,agrawal1998mining}\inserted{\citep{van2004workflow} 
    with applications in bioinformatics \citep{yan2005mining}, social-network analysis \citep{carrington2005models} or web navigation \citep{papadimitriou2010web}}. However, we take advantage of the properties derived from the graphs to propose an inductive method that is able to generate more than one model, each one identifying a different form of performing the task showed by the examples.

    
    \item The method is driven by two main operators: aggregation and 
    refinement. Those operators perform the generalisation of the examples (by aggregating graphs) and the refinement of the aggregated graph  (by removing some edges according to a certain threshold). Unlike other process mining methods based on graphs, our method applies the two operators repeatedly generating one model in each iteration,  until  all the training examples are represented by at least one of the inferred models.
    

    \item The learned models are characterised by not having 
    disconnected nodes, so the ending activity is always reachable from the initial activity following a path in the graph; in other words, the models represent correct ways to perform the task.  
    
    
    \item While other methods for learning tasks from a few executions get black-box models that are not comprehensible or their decisions are impossible to explain  \citep{duan2017one}, our models are  completely understandable by the experts since there is a one-to-one correspondence between nodes in the model and activities. Hence, the whole learning process can be audited.
    
    \end{itemize}

\noindent    
\modified{We have applied our method to two real problems. The first one  is a suturing task from the domain of skill training in minimally invasive surgery (MIS), which we will use as a running example. This is one of the most complicated routines to perform because of  noise \citep{cao1996task} and because surgeons can express different suture styles \citep{ahmidi2017dataset}.}\inserted{ The second real application is related to cooking: the preparation of a brownie \citep{spriggs2009temporal}. This allows us to analyse the applicability of the model to very different domains.} 

The rest of this paper is organised as follows.  
\inserted{Section \ref{related} reviews some  related work.} Section \ref{matandmet} presents the notation we use for representing examples and models. We also define  our inductive method for learning multiple models of a task.
\modified{Section \ref{s:results} presents the experimental results of the application of our approach to real examples taken from training in two domains: laparoscopic surgery and cooking.} In Section \ref{s:discussion} we discuss the experimental results and the strengths and limitations of our method. 
Finally, Section \ref{s:conclusions} outlines the conclusions and further research directions.

\section{\inserted{Related Work}}\label{related}

\inserted{Traditionally, the recognition of complex human activities from a set of 
observed data has been addressed in the area of Activity Recognition (AR) \citep{sukthankar2014plan}. In AR, a distinction can be made between data-driven approaches and knowledge-driven approaches \citep{chen2012sensor}. }

\inserted{Data-driven approaches \citep{hoey2010automated,sanchez2008activity,patterson2005fine,kruger2014computational} are characterised by the use of supervised (e.g., Hidden Markov Models (HMM), Linear Dynamical 
Systems (LDSs)) and unsupervised (e.g., KNN) machine learning techniques to address the problem. These are powerful tools when facing uncertainty and temporal information but they require large datasets for learning the activities. By contrast, knowledge-driven approaches \citep{okeyo2011ontology,ye2015usmart,chen2012knowledge,chen2008logical,bouchard2006smart} are characterised by their reusability, semantic clarity and a lower dependence on the training data. Activity models consist of rules that define the logic of the task and its constraints. This representation makes it possible to introduce domain knowledge. 
This prior knowledge can be  easily translated into reusable structures, i.e., schemes, logical representation or ontologies, which are then used for reasoning about the relationships between activities, objects, temporal and spatial context. }

The applicability of both types of approaches inside a real automatic supervision system has become a hot topic in recent years. Thus, data-based systems have been proposed, for instance, for assisting persons with dementia during handwashing \citep{hoey2010automated}, for kitchen activities supervision  \citep{kruger2014computational, yordanova2017s,rohrbach2012database,neumann2017kognichef} and for supervision in a whole smart environment using sensor data \citep{twomey2016sphere,intille2006using,krose2008care,crandall2011tracking}. Additionally, knowledge-based supervision systems try to take full advantage of other knowledge resources in contexts where accessing training data is complicated. For instance, in \citep{okeyo2011ontology,ye2015usmart,chen2012knowledge} different knowledge-driven approaches are proposed to recognise the behaviour of smart home inhabitants by using ontologies. \citet{chen2008logical} 
present a knowledge-driven framework for Smart Homes based on the Event Calculus (EC) formalism \citep{mueller2014commonsense}, and \citet{bouchard2006smart} present a similar idea to prevent home accidents of people that suffer from cognitive impairments (e.g., Alzheimer). The dependence on data makes data-based systems  sensitive to issues such as data scarcity \insertedd{(it is difficult and costly to have enough training data from different users, specially when data is collected from devices such as sensors)} or even the ``cold start'' problem: the problem that emerges when a data-driven application does not have the minimum amount of data to learn the task and operate. In addition, limitations in scalability and reusability may also arise  due to difficulties in applying the learnt models from one person to another \insertedd{(since humans perform activities differently, they have different activity patterns). A way to deal with those problems is to apply multitask learning \citep{caruana1997multitask}  by considering  each user as a task \citep{liu2015action2activity,sun2012large} or considering each activity as a task and solving multiple tasks by  exploiting  the  commonalities  and  differences  across them \citep{peng2018aroma}. An alternative is to perform transfer-based activity recognition  \citep{cook2013transfer} by transferring the knowledge learnt from source domains (users, body parts, devices, etc.) to a target domain \citep{van2010transferring,chen2019cross,ding2019empirical}.} 
\insertedd{Above all, despite being possible to describe a scene by recognising the sequence of actions, in many cases, this is meaningless for understanding the task by itself, so a high-level reasoning expressing the domain knowledge is required to those effects}.
On the other hand, knowledge-based  systems have limitations when handling noisy and uncertain data, as well as managing temporal information. In some cases, these models could be viewed as quite rigid and incomplete \citep{chen2012sensor}.

\inserted{The issue of determining the process behind a human task has been tackled 
in many ways. 
An interesting line of work is to create a completely connected model and then learn the transitions from a set of training runs. Probabilistic models are often used, such as Hidden Markov Models (HMMs) \citep{boger2005decision,kalra2013detection,duong2009efficient} or Dynamic and Naïve Bayes networks \citep{baker2009action,dai2008group,oh2014probabilistic}. The learnt model is better adapted to the real behaviour of the process since it is based on evidence.  However it suffers from the same drawbacks of data-based approaches since a large evidence is required to learn the transition probabilities. Additionally, in 
complex domains 
the model can end up in an explosion of possible states increasing the computational cost of these approaches  considerably 
\citep{sadilek2012location,rosen2002task}.}

\inserted{Another two areas that address the challenge of inferring a process model directly from observing demonstrative examples are imitation learning  and process mining. The objective of imitation learning \citep{hussein2017imitation} is that a machine (robots in many cases) can learn the necessary movements (i.e., policy) to solve a simple task through examples performed by a human expert while the machine observes it. The current trend in imitation learning focuses on the use of deep learning techniques \citep{xu2018neural,duan2017one,finn2017one}. Although they have given surprising results, the tasks they are able to learn and solve are still very basic (e.g., stacking boxes).  In addition, the fact of using deep learning hinders the traceability and analysis of the knowledge acquired by the system. On the other hand, process mining  \citep{agrawal1998mining,van2016process} includes a family of techniques that support the analysis of business processes 
from some observations on how they are currently being executed. More concretely, process discovery focuses on learning  process models from traces of activities (i.e., real process executions or  \textit{event log}). The generated model represents the main flow of activities in the  process at hand. Process mining has been mainly applied to business management \citep{van2007business} but also to other fields such as healthcare \citep{mans2008application,yang2017medical}. Although the previous techniques are able to explain the processes in an organisation, they cannot be directly applied for the purpose of supervision: the models may contain non-essential activities for carrying out the process and they are not able to capture the different ways (if any) of performing a process since only one model is built.}

\insertedd{In this paper we cover two application domains:  skill training in minimally invasive surgery and cooking. We believe these two domains are representative of situations where there are more than one possible way of performing a task, either because of symmetries (left and right hands in surgery) or indistinct sequences (ingredient order).  
Other datasets in the cooking domain, such as the salad preparation dataset \citep{stein2013combining} or the sandwich preparation dataset \citep{spriggs2009temporal}, present a huge variability in executions, not really displaying a short number of patterns, and applying pattern extraction to them would not be very meaningful (probably ending up with a pattern per example). Our method is designed for  well-structured high-to-medium level activity sequences. This occurs in tasks such as those described in the case studies:  surgery and recipes.}

\insertedd{There are domains that may show one or a small number of patterns, but they are usually found in the area of activity recognition  \citep{van2011human,voulodimos2012threefold,liu2015action2activity}. 
These datasets usually contain low-level data extracted directly from devices such as sensors or smartphones. Therefore, in order to apply our method to these domains we would need to add a first phase to extract and determine the events, and then a second phase (using the proposed algorithm) to learn the sequence of events. The quality of the final process mining (and the number of patterns found) would be strongly determined by the framing of the activity recognition stage (and the choice of method there) and not only by our method, making the comparison of results prone to too many confounding factors. Nevertheless, beyond the scope of the paper, we see a lot of potential in the future to explore and apply our method in the best possible combinations with methods for activity recognition. 
}

\section{A graph-based method for inferring Multiple Models of a Task}
\label{matandmet}
In this section, we present a method to learn the different ways of carrying out a task from  several executions of the task performed by experts.  We consider a task execution as a sequence of activities that are realised consecutively with a start and an end. Following that sequence, the task is successfully completed. Formally, we define $\mathcal{A}$ as the set of activities (vocabulary) that can be used to accomplish the task, being an execution example a finite sequence of these activities $\delta = (a_1, a_2, \ldots, a_n), a_i \in \mathcal{A}$. We denote the full set of sequences provided to the system as $\Delta$. 

Working directly with sequences has several limitations in order to learn and express, in an intuitive and simple way, how a task should be realised. Firstly, 
if the sequence is too long, it becomes extremely difficult to understand the flow of activities that is taking place. Note that the main aim is to obtain a  model of the task that could be used for training and/or supervising non-expert apprentices. Therefore, 
it is crucial that the learnt model can be easily interpreted by a human. Secondly, the problem of learning from task executions (especially when the tasks are complicated to execute because of the skills they require) is that the sequences may contain infrequent activities or noise. In our case, noise are those activities that are not really needed to complete the task\footnote{In other approximations, noise is considered any missing activity or any non-executed activity inserted in a sequence $\delta$. However, the treatment of this kind of noise is out of the scope of the problem discussed in this paper.}. It may be complicated to detect during the learning process whether the  elimination of these noisy activities from the sequences leads to invalid models that are not able to complete the task, 
because the same activity can be essential in one part of the task and unnecessary in another part.
Therefore, 
the use of a graph-based representation language to compactly simplify the sequences of activities as dependency graphs helps in this sense (see Section \ref{ss:plan_as_multigraph}). In this case, the possible loss of information due to the change of representation is clearly compensated because dependency graphs  greatly facilitate the process of identifying which activities are essential to perform a task (and therefore will be part of the models), which activities are dispensable because they are non-essential, as well as the formal confirmation that a model really express a correct way to fulfil the task.

\begin{figure}[ht!]
\centering
\captionsetup{justification=centering,margin=2cm}
  \includegraphics[width=1.0\columnwidth]{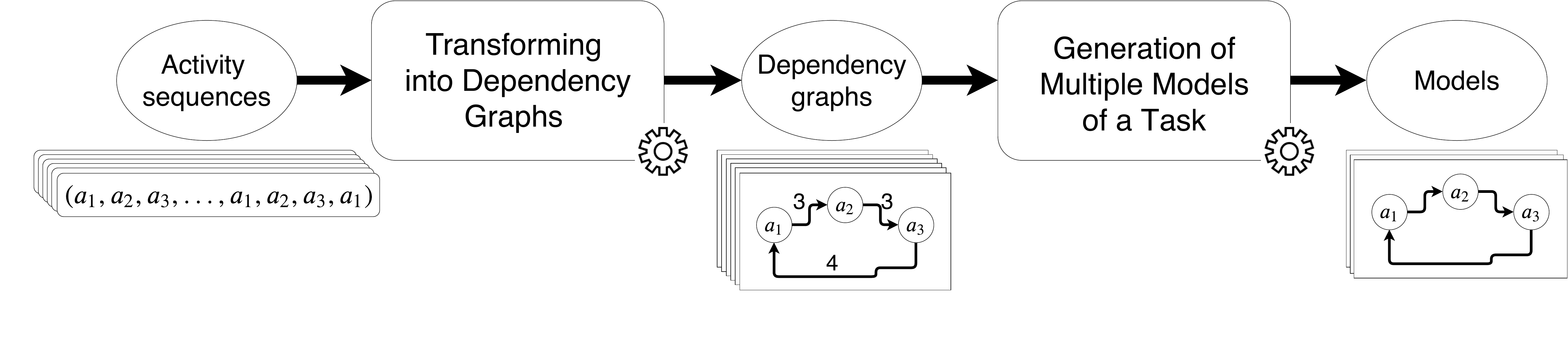} 
  \caption{The pipeline to learn the task models.}~\label{fig:pipeline_approach}
\end{figure}

The general pipeline 
of our approach is shown in Figure \ref{fig:pipeline_approach}. In the first step, 
each sequence of activities is simplify into a dependency graph.  
The second step is the  algorithm that infers the different models. 

\subsection{Graph-based  formulation: notation and definitions}\label{ss:notation}
A dependency graph is a labelled directed graph $G = (V, E)$, where $V$ is a set of labelled 
vertices, and $E \subseteq V \times V$ is a set of weighted directed edges, such that each edge has a weight given by a weight function $\omega: E \rightarrow \mathbb{N}_{\geq 1}$. 
The labels of the vertices belong to a finite set of labels $\mathcal{L} = \{l_1,\ldots,l_n\}$, where each $l_i$, $1\leq i \leq n$, denotes an activity  $a_i \in \mathcal{A}$. $V$ contains two special vertices, $v_S$ and $v_F$, that represent two synthetic activities denoting the starting and ending of the task. The two synthetic activities are implicitly placed at the beginning and end of each activity sequence, respectively. 
Analogously, the set of labels $\mathcal{L}$ is extended to include two special labels $S$ and $F$ associated to the vertices $v_S$ and $v_F$, respectively.
For the sake of readability, given a graph $G$, $V(G)$ and $E(G)$ denote the vertex and edge sets of $G$, and  $\omega_{G}$ denotes its weight function. 
Given an edge $e=(v_i, v_j)$, $v_i$ and $v_j$ are the source and target vertices of $e$, which can be retrieved using the functions $\sigma$ and $\tau$, respectively (i.e., $\sigma(e)=v_i$ and $\tau(e)=v_j$). 
Given a set of graphs $D = \{G_1, G_2, \ldots, G_n\}$ and an edge $e$, we define $D_e$ as the subset of $D$ that contains $e$, that is, $D_e = \{G_i \in D \:|\: e\in E(G_i)\}$.










In what follows we introduce the notions of walk 
and validity, and three types of distinctive graphs that are used by the learning algorithm.

\theoremstyle{definition}
\begin{definition}{\scshape (Walk).}\label{def:path_graph}
Given a dependency graph $G = (V, E)$, and two vertices $\{v_i, v_j\} \in V$, a walk $w(v_i, v_j)$ between $v_i$ and $v_j$ is any sequence of directed edges $(e_1, e_2 \ldots, e_n)$ from $E$, such that $\sigma(e_1)=v_i$, $\tau(e_n)=v_j$, and for any pair of consecutive edges $e_k$ and $e_{k+1}$, $\tau(e_k)=\sigma(e_{k+1})$. A walk between $v_S$ and $v_F$ is called a complete walk, denoted as $\hat{w}$. 
\end{definition}

\theoremstyle{definition}
\begin{definition}{\scshape (Validity).}\label{def:validity}
Given a 
dependency graph $G = (V, E)$, we say that $G$ is valid if $V_S = V_F$, where  $V_S =\{ v_i \in V \setminus \{v_S,v_F\} \;|\; \exists w(v_S, v_i)\}$ and $V_F =\{ v_i \in V \setminus \{v_S,v_F\} \;|\; \exists w(v_i, v_F)\}$.\end{definition}

\theoremstyle{definition}
\begin{definition}{\scshape (Aggregated graph).}\label{def:aggregation}
Given a set of dependency graphs $D = \{G_1, G_2,$ $\ldots, G_n\}$, the aggregation of $D$ is the graph $\AggregatedGraph = (\AggregatedGraphVertex{}, \AggregatedGraphEdge{})$, 
such that $\AggregatedGraphVertex{} = \bigcup_{G_i \in D} V(G_i)$, $\AggregatedGraphEdge{} = \bigcup_{G_i \in D} E(G_i)$, and $\forall e \in \AggregatedGraphEdge{}, \omega_{\AggregatedGraph{}}(e) = \sum_{G_i \in D_e} 
\omega_{G_i}(e)$. We call $\AggregatedGraph$ as the \textit{aggregated  graph}. 
\end{definition} 



\theoremstyle{definition}
\begin{definition}{\scshape (Intersected graph).}\label{def:intersection}
Let $G_1 = (V_1, E_1)$ and $G_2 = (V_2, E_2)$ be two 
dependency graphs, the intersection of $G_1$ and $G_2$ is defined as  $\IntersectedGraph = G_1 \cap G_2= (V_1 \cap V_2, E_1 \cap E_2)$, where $\forall e \in E_1 \cap E_2$, $\omega_{G^\cap{}}(e)=min(\omega_{G_1}(e),\omega_{G_2}(e))$.
\end{definition}

\theoremstyle{definition}
\begin{definition}{\scshape (Threshold graph).}\label{def:thresholding}
Given a 
dependency graph $G = (V, E)$ and a threshold 
$\theta \in \mathbb{N}_{\geq 1}$, we define the \textit{threshold graph} $\ThresholdGraph = (V_\theta, \ThresholdEdges)$ as the subgraph of $G$ such that  $\ThresholdEdges = \{e \in E \:|\: \omega_{G}(e) \geq \theta\}$ and $V_\theta = \{\sigma(e) \:|\: e \in \ThresholdEdges\} \cup \{\tau(e) \:|\: e \in \ThresholdEdges\}$.
\end{definition}

\theoremstyle{definition}
\begin{definition}{\scshape (Overlap).}\label{def:coverage}
Given two dependency graphs $G_1$ and $G_2$, we say that $G_1$ and $G_2$ overlap if their intersection $\IntersectedGraph = G_1 \cap G_2$ is valid. 
\end{definition}



\theoremstyle{definition}
\begin{rmk} 
From the above definitions we remark that: (1) 
there are complete walks in a valid dependency graph; (2) if an intersected graph $\IntersectedGraph= G_1 \cap G_2$ is valid then it contains all the common complete walks of $G_1$ and $G_2$;  (3) 
the application of a threshold 
does not preserve the graph validity property.  
\end{rmk}

In what follows we use the labels of the vertices to denote them. 

\subsection{Data formalisation: from activity sequences to dependency graphs} \label{ss:plan_as_multigraph}

\begin{figure}[t!]
\centering
  \includegraphics[width=0.8\columnwidth]{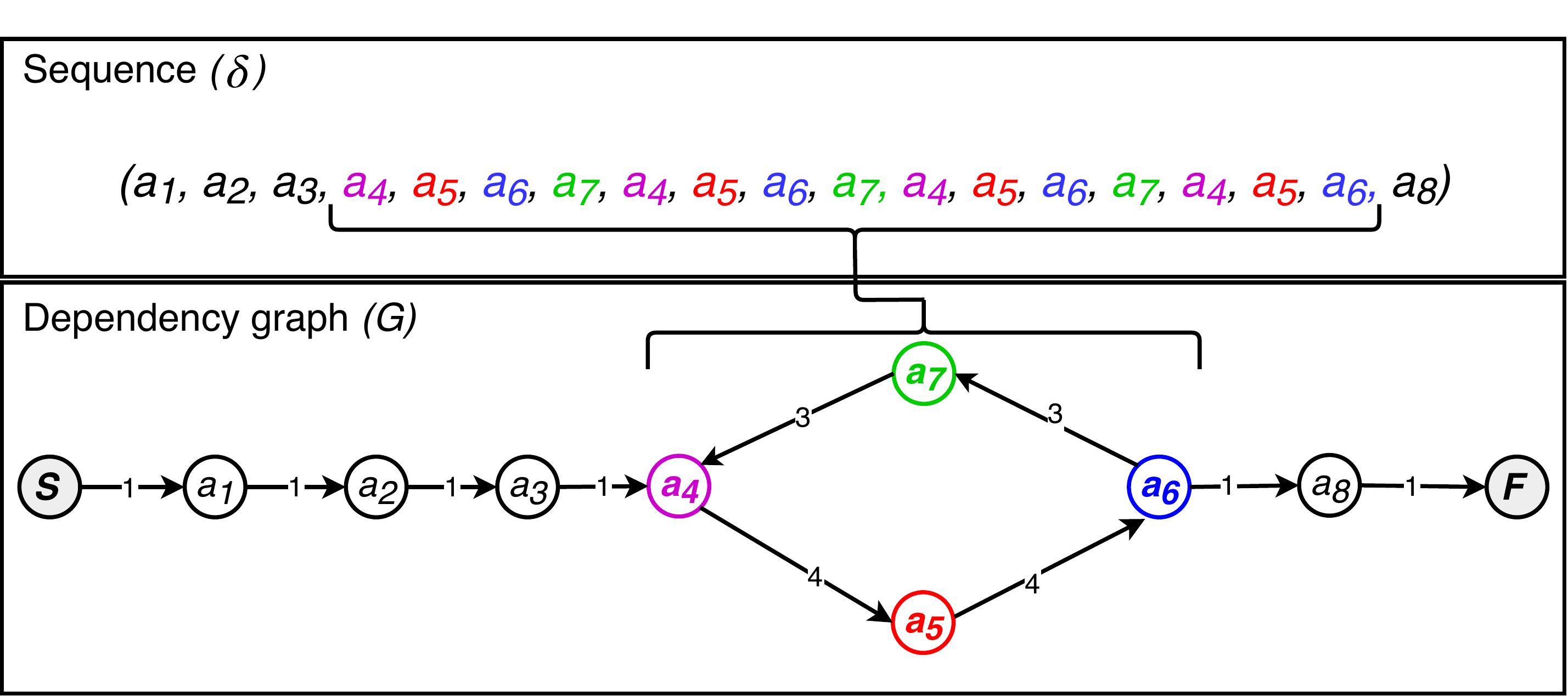} 
  \caption{A sequence of activities (on the top) expressed as a dependency graph  (on the bottom). The vertices corresponding to the synthetic activities (S and F) are highlighted in bold and shadowed, and the activities that appear more than once are depicted in colour. 
  }~\label{fig:sequence_to_graph}
\end{figure}

In this section, we describe how to convert an 
activity sequence  $\delta = (a_1,\ldots, a_n)$ into a dependency graph $G$.  
Firstly, $V(G)$ is created containing $v_S$, $v_F$ and as many vertices $v_i$ as different activities $a_i$ are in $\delta$ with labels $l_i=a_i$. $E(G)$ is also initialised by containing the directed edges $(S,a_1)$ and $(a_n,F)$ with weights equal to $1$. 
Then, for each pair of consecutive activities  $(a_i,a_j)\in \delta$, if the edge $e=(a_i,a_j)$  already belongs to $E(G)$, then its weight is increased $\omega_G(e)= \omega_G(e)+1$; otherwise, $e$ is added to $E(G)$ with weight $\omega_G(e)=1$. Figure \ref{fig:sequence_to_graph} illustrates this conversion process. On the top, it is shown an activity sequence of length 19 composed by 8 different activities ($a_1$ to $a_8$), some of them appearing more than once (marked in colour). This sequence is expressed 
as the dependency graph on the bottom of this figure, which is formed by 10 vertices and 10 edges. The labels on the edges indicate their weights. For instance, the weight of edge $(a_4, a_5)$ is $4$ because the activity $a_4$ appears followed by the activity $a_5$ four times in the sequence.

Note that a dependency graph 
is a representation more concise than the original activity sequence. That means that different sequences can produce the same dependency graph. This can happen when there are activities that are performed more than once along the sequence.  
 This could imply losing some information about the order in that the activities have been done when the sequence turns into a dependency graph.  For instance, the sequences $\delta_1 = (a_1,a_2,a_3,a_4,a_2,a_4,a_2,a_4,a_5)$ and $\delta_2 = (a_1,a_2,a_4,a_2,a_4,a_2,a_3,a_4,a_5)$ generate the same dependency graph. Thus, by looking at the graph we can only say that activity $a_3$ is performed after activity $a_2$ but not when (after doing $a_2$ for the first, the second or the third time). This fact is known 
 as ``representational bias'' in the area of process mining   (\cite{van2016process}) or ``language bias'' in the inductive logic programming field (\cite{ade1995declarative}), and it refers to the implicit choices that restrict the syntax of examples and hypothesis.  By contrast, one of the main benefits of the language bias is that it reduces  
 the search space of possible models. In our case, the graph representation allows us to search the 
 models by systematically applying the graph operations defined in Section \ref{ss:notation}.

\subsection{Mining Multiple Models from dependency graphs} \label{ss:mining_process}

In this section we describe a new method for inducing multiple models from a set of dependency graphs $\mathcal{G} = \{G_1, G_2, \ldots, G_n\}$. A model $M$ is a dependency graph that satisfies the following conditions:
\begin{itemize}
    \item $M$ is valid and, consequently, there exists  in $M$ (at least) one walk from $S$ to $F$. In terms of the original task, this means that such a walk indicates the activities that are needed to perform the task.
    \item $M$ has to overlap at least one dependency graph in $\mathcal{G}$. Newly, in terms of the original task, this means that there exits at least one expert execution that has performed the same activities included in $M$. In other words, $M$ captures one of the ways of solving the task according to the experts. 
  
\end{itemize}
Additionally,  the presence of noise in $M$ (non-essential activities for the task) should be minimised as much as possible, in order to obtain the desired trade-off between generality and specificity.

The {\scshape\ShortNameMiningAlgorithm{}} algorithm (Algorithm \ref{alg:PM_algorithm}) combines a generalisation operator (graph aggregation) with a refinement operator.  After applying both operators, the dependency graphs that overlap (Definition \ref{def:coverage}) with the induced model are removed. This process is repeated using the remaining  graphs until all the dependency graphs  overlap with one of the models. In this way, we are able to infer the different styles of solving the task showed by the experts in their executions. The {\scshape\ShortNameMiningAlgorithm{}} algorithm is inspired by sequential covering strategies of rule learning \citep{furnkranz1999separate}. 


\begin{algorithm}
\begin{algorithmic}[1]
\REQUIRE a set of dependency graph: $\mathcal{G} = \{G_1, G_2, \ldots, G_n\}$.
\ENSURE a set of models: $\mathcal{M}$. 
  \STATE $\mathcal{M} \leftarrow \emptyset$. \label{lst:PM:line:1}
  \STATE // Iterate over the set of graphs until each $G_i$ has a $M$ 
  that overlaps with it.
 \WHILE{$\mathcal{G} \neq \emptyset$}\label{lst:PM:line:2}
  	\STATE $\AggregatedGraph \leftarrow$ AggregateGraphs($\mathcal{G}$). \label{lst:PM:line:3}
    \STATE // $\bar{\mathcal{G}}$ is the set of overlapped dependency graphs.  
    \STATE \{$M$, $\bar{\mathcal{G}}$\} $\leftarrow$ {\scshape\FullNameThresholdingAlgorithm{}}($\AggregatedGraph$, $\infty$, $\mathcal{G}$). \label{lst:PM:line:4} 
     \STATE $\mathcal{G} \leftarrow \mathcal{G} \setminus \bar{\mathcal{G}}$. \label{lst:PM:line:5}
     \STATE $\mathcal{M} \leftarrow \mathcal{M} \cup \{M\}$. \label{lst:PM:line:6}
  \ENDWHILE \label{lst:PM:line:7}
  \RETURN $\mathcal{M}$ \label{lst:PM:line:8}
\end{algorithmic}
\caption{\scshape\ShortNameMiningAlgorithm{}  Algorithm}\label{alg:PM_algorithm}
\end{algorithm}

More concretely, the {\scshape\ShortNameMiningAlgorithm{}} algorithm works as follows. 
First, the most general graph is generated by aggregation (Line \ref{lst:PM:line:3}). This aggregated graph contain all the vertices and edges included in the dependency graphs $G_i \in \mathcal{G}$. Clearly, this aggregated graph is too much general: (1) it contains not only valid walks but other new walks (not included in any $G_i$) by combining partial walks from several $G_i$ (that probably do not represent a correct way to execute the task), and (2) it is the noisiest graph of all that can be generated from $\mathcal{G}$. Hence, $\AggregatedGraph$ must  be  refined below (Line \ref{lst:PM:line:4}). This refinement returns one model $M$ and the set of dependency graphs $\bar{\mathcal{G}}$ that overlap with $M$. Then, the set of dependency graphs $\bar{\mathcal{G}}$ that overlap with the refined graph $M$ are removed from $\mathcal{G}$ (Line \ref{lst:PM:line:5})  and $M$ is added as a part of the solution (Line \ref{lst:PM:line:6}). Then, the next iteration of the algorithm starts with the remaining examples, creating a new aggregated graph and so on. This process is repeated until complete the solution.

\begin{figure}[ht!]
\centering
  \includegraphics[width=0.7\columnwidth]{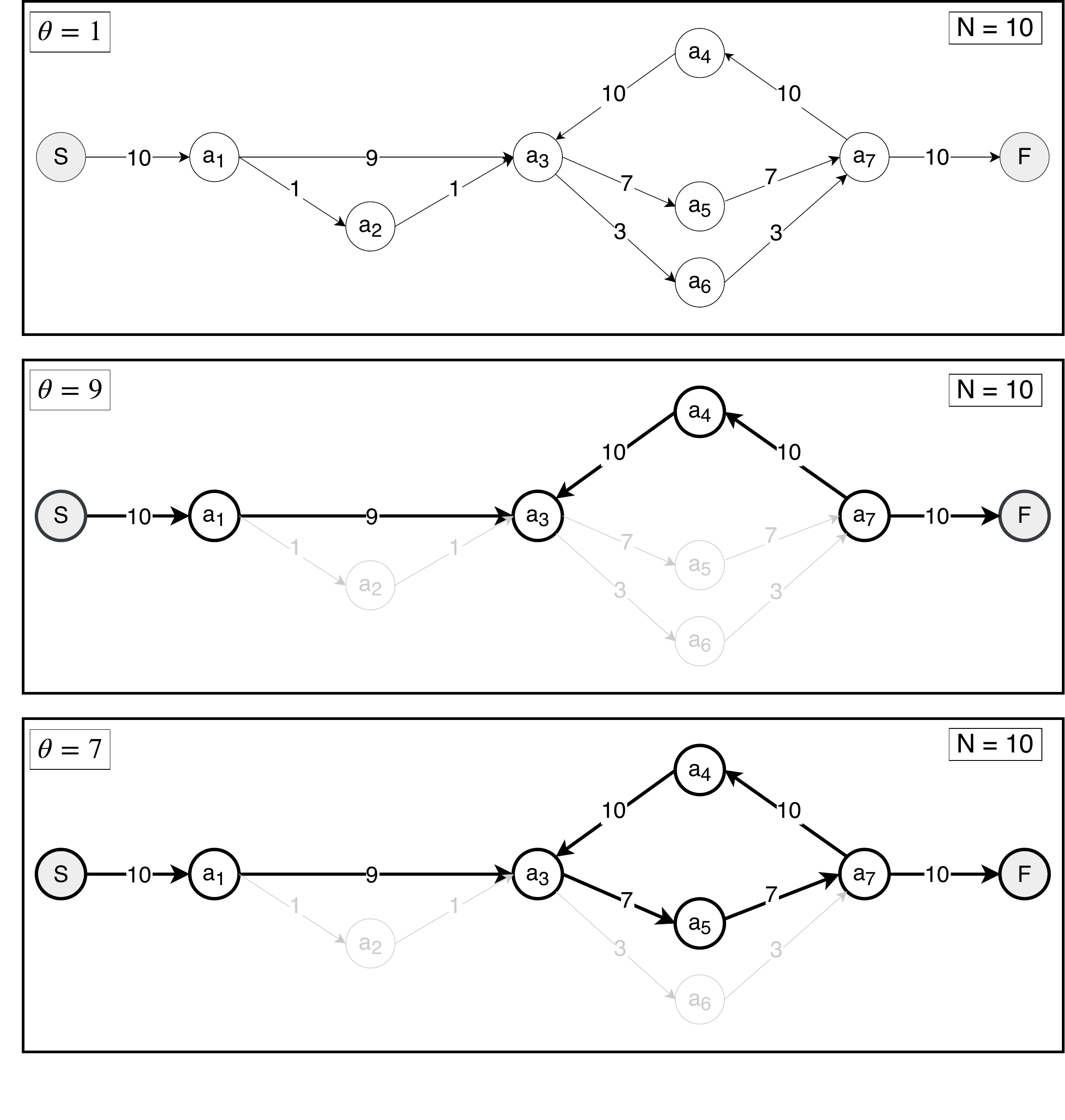} 
  \caption{Effect of applying different thresholds for refining a $\AggregatedGraph$ formed by  the aggregation of 10 dependency graphs (shown on the top). Note that 
  the refinement  procedure keeps all the edges of $\AggregatedGraph$ whose weight is greater than $\theta$.}~\label{fig:thresholding_example} 
\end{figure}

The refinement of the aggregated graph $\AggregatedGraph$ (Algorithm \ref{alg:T_algorithm}) is performed by repeatedly applying a threshold $\theta$ to $\AggregatedGraph$, that is, by removing edges according to their weights.  
Hence, we vary the threshold decreasingly starting from a maximum value, so the first graphs $\ThresholdGraph[+]$ that are generated are usually very simple as many edges are removed. The application of a threshold has the aim to remove the non-essential activities from $\AggregatedGraph$. Progressively, we reduce the threshold and apply it until get a  $\ThresholdGraph[+]$ that satisfies the conditions to be a model: it is valid and overlaps at least with a dependency graph $G_i$. Figure \ref{fig:thresholding_example} illustrates  the application of a threshold to an aggregated graph. On the top, we see an aggregated graph $G^+$ formed by $N=10$ dependency graph. When we use a high threshold (i.e., $\theta=9$), the resulting threshold graph (shown in the middle of the figure) is not  valid. 
Meanwhile, by taking a lower threshold (for instance, $\theta=7$), we obtain a valid threshold graph (shown on the bottom) that represents a way to completely perform the task. It is worth noting that some activities that are rarely performed in these examples (denoted by the low weight of their incoming and outgoing edges in $G^+$), such as $a_2$ or $a_6$, are not included in $G^+_{\theta=7}$.

The input of the {\scshape \FullNameThresholdingAlgorithm{}} 
algorithm (Algorithm \ref{alg:T_algorithm})  is  $\AggregatedGraph$, an initial threshold $\theta$,  and $\mathcal{G}$ (the set of dependency graphs used to generate $\AggregatedGraph$).
In the case  the initial threshold is $\infty$, the algorithm sets the $\theta$ value at the minimum weight that guaranties that  $\ThresholdGraph[+]$ could be valid (Lines \ref{lst:T:line:2}-\ref{lst:T:line:8}). 
The $\theta$ value 
is firstly used on $G^+$ to discard those edges whose weight is below that value (Line \ref{lst:T:line:9}). 
There are two reasons why this $\theta$ may be inappropriate to generate a model: (1) it produces an invalid  $\ThresholdGraph[+]$,  or (2) $\ThresholdGraph[+]$ is valid but does not overlap with any $G_i$. In the first  case, the $\theta$ value is repeatedly decreased\footnote{For efficiency reasons, the next value for $\theta$ is selected from the weights of the edges in $E^+$, assuming  $E^+$ ordered in decreasing order of their edges weights.} 
until a valid $\ThresholdGraph[+]$ is found\footnote{Observe that selecting the $\theta$ values in decreasing order of the weights in $E^+$ assures  the termination of the  algorithm \ref{alg:T_algorithm}, since in the worst case the minimum weight in $E^+$ is taken as $\theta$, which means that no edges are removed from the aggregated graph, and therefore $\ThresholdGraph[+]$ is a valid graph that overlaps with all the dependency graphs used to construct it.} (Lines \ref{lst:T:line:10}-\ref{lst:T:line:13}). 
For case (2), the next lower $\theta$ value is selected 
as in the first case, and  
a recursive call is performed (Lines \ref{lst:T:line:20}-\ref{lst:T:line:23}). 
 According to Definition \ref{def:coverage}, to check whether a threshold graph $\ThresholdGraph[+]$ overlaps with a dependency graph $G_i$,  we verify the validity of their intersection $G_i \cap \ThresholdGraph[+]$  
 (Lines \ref{lst:T:line:14}-\ref{lst:T:line:19}). 

Finally, when a valid $\ThresholdGraph[+]$ overlaps with one or more examples, the algorithm returns $\ThresholdGraph[+]$ as a model, and the list of $G_i$ that overlap with it (Line \ref{lst:T:line:24}).

\begin{figure}[ht!]
\centering
  \includegraphics[width=0.8\columnwidth]{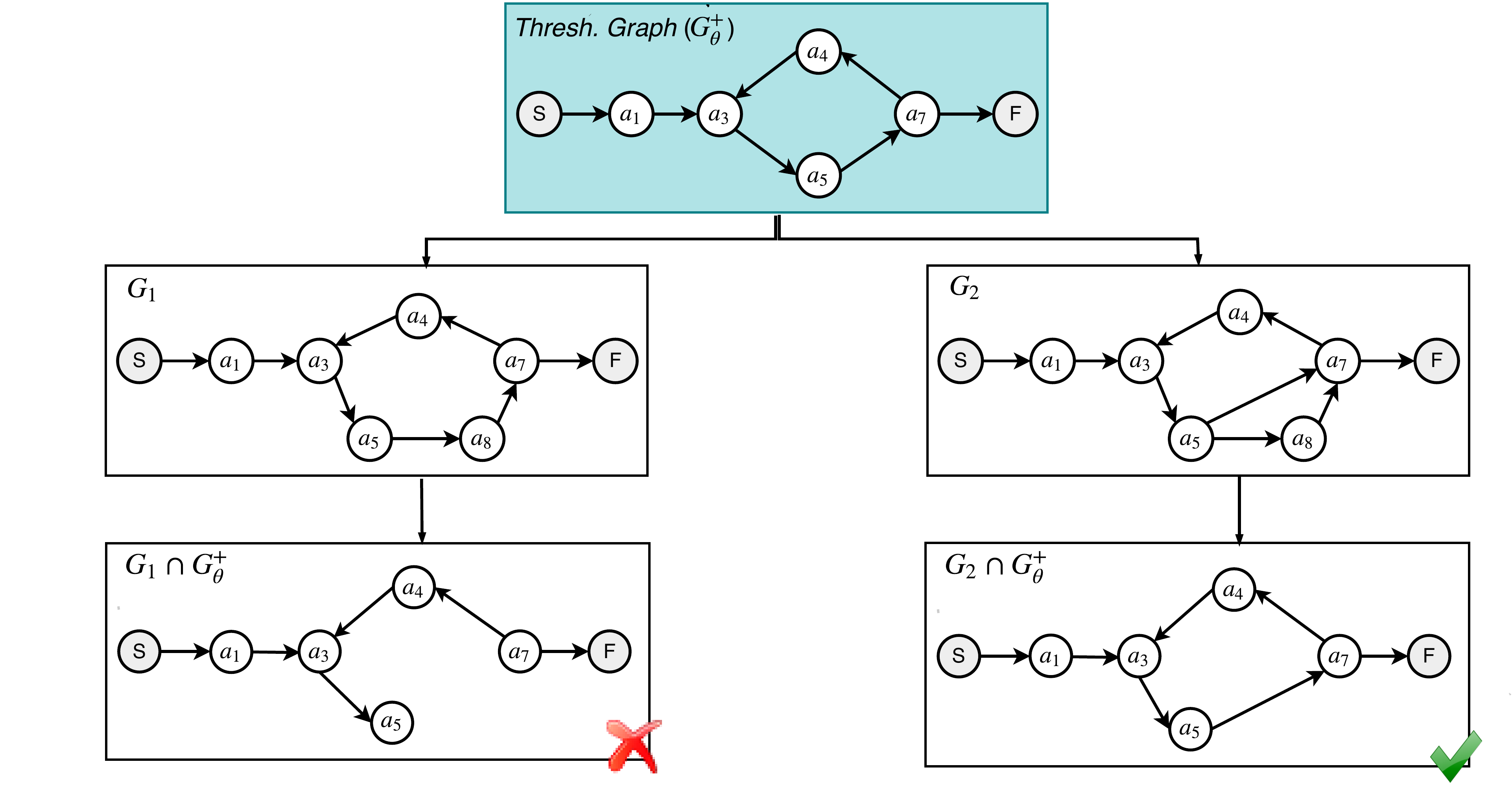} 
  \caption{ Overlap checking between a threshold graph $\ThresholdGraph[+]$ and two dependency graphs $G_1$ and $G_2$. There is no overlapping between $\ThresholdGraph[+]$ and $G_1$ because their intersection $G_1 \cap \ThresholdGraph[+]$ (on the left bottom) does not hold the validity constraint ($V_S \neq V_F$). In contrast, $\ThresholdGraph[+]$ overlaps with $G_2$ because their intersected graph $G_2 \cap \ThresholdGraph[+]$ (on the right bottom) is valid ($V_S = V_F$).}~\label{fig:intersection_example}
\end{figure}

Figure \ref{fig:intersection_example} shows an example of how the overlap between graphs is checked. From top to bottom, we see one threshold graph $\ThresholdGraph[+]$, two dependency graphs  $G_1$ and $G_2$, and the intersected graphs $G_1 \cap \ThresholdGraph[+]$ (left) and $G_2 \cap \ThresholdGraph[+]$ (right). Note that $G_1 \cap \ThresholdGraph[+]$ is not a valid dependency graph because it does not contain any complete walk, which means that $\ThresholdGraph[+]$ does not capture the way of performing the task showed by the complete walk in $G_1$.  However, the sets $V_S$ and $V_F$ in $G_2 \cap \ThresholdGraph[+]$ are equal, $V_S = V_F = \{a_1, a_3, a_5, a_8, a_7, a_4\}$, 
and hence  $\ThresholdGraph[+]$ overlaps with  $G_2$.

\begin{algorithm}
\algsetup{linenosize=\tiny}
\small 
\begin{algorithmic}[1]
\REQUIRE an aggregated graph: $\AggregatedGraph = (V^+, E^+)$; a threshold value: $\theta$; a set of dependency graph: $\mathcal{G} = \{G_1, G_2, \ldots, G_n\}$.
\ENSURE a model $M$: $\ThresholdGraph[+]$; a set of overlapped examples: $\bar{\mathcal{G}}$. 
    \STATE $\bar{\mathcal{G}} \leftarrow \emptyset$. \label{lst:T:line:1}
    \STATE // Find the minimum weight to begin and end a $\hat{w}$.
    \IF{$\theta$ is $\infty$}  \label{lst:T:line:2} 
      \STATE $v_S \leftarrow$ GetStartVertex($V^+$).  \label{lst:T:line:3}
      \STATE $v_F \leftarrow$ GetEndVertex($V^+$).  \label{lst:T:line:4}
      \STATE $starting\_weight \leftarrow$ MaxEdgeWeight($E^+$, $v_S$).  \label{lst:T:line:5}
      \STATE $ending\_weight \leftarrow$ MaxEdgeWeight($E^+$, $v_F$). \label{lst:T:line:6}
      \STATE $\theta \leftarrow$ min($starting\_weight$, $ending\_weight$). \label{lst:T:line:7}
   \ENDIF \label{lst:T:line:8}
   \STATE $\ThresholdGraph[+] \leftarrow$ ApplyThreshold($\AggregatedGraph$, $\theta$). \label{lst:T:line:9}
   \STATE // Adjust the $\theta$ value in order to get a valid $\ThresholdGraph[+]$.
   \WHILE{not IsValid($\ThresholdGraph[+]$)} \label{lst:T:line:10}
     	\STATE $\theta \leftarrow$ NextWeightBelowThreshold($E^+$, $\theta$). \label{lst:T:line:11}
		\STATE $\ThresholdGraph[+] \leftarrow$ ApplyThreshold($\AggregatedGraph$, $\theta$). \label{lst:T:line:12}
   \ENDWHILE \label{lst:T:line:13}
   \STATE // Check whether the valid $\ThresholdGraph[+]$ also overlaps at least one $G \in \mathcal{G}$.
   \FOR{each $G$ in $\mathcal{G}$}  \label{lst:T:line:14}
     \STATE $\IntersectedGraph \leftarrow$ $G \cap \ThresholdGraph[+]$. \label{lst:T:line:15}
     \IF{IsValid($\IntersectedGraph$)}    \label{lst:T:line:16}
       \STATE $\bar{\mathcal{G}} \leftarrow \bar{\mathcal{G}} \cup G$. \label{lst:T:line:17}
     \ENDIF \label{lst:T:line:18}
   \ENDFOR \label{lst:T:line:19}
    \STATE // Do a recursive call with a lower $\theta$ if any $G$  overlaps with the current $\ThresholdGraph[+]$.
    \IF{$\mathcal{\bar{G}}$ is $\emptyset$} \label{lst:T:line:20}
      \STATE $\theta \leftarrow$ NextWeightBelowThreshold($E^+$, $\theta$). \label{lst:T:line:21}
      \STATE \{$\ThresholdGraph[+]$, $\mathcal{\bar{G}}$\} $\leftarrow$ {\scshape\FullNameThresholdingAlgorithm{}}($\AggregatedGraph$, $\theta$, $\mathcal{G}$).  \label{lst:T:line:22}
    \ENDIF \label{lst:T:line:23}
	\RETURN \{$\ThresholdGraph[+]$, $\bar{\mathcal{G}}$\} \label{lst:T:line:24}
\end{algorithmic}
\caption{{\scshape\FullNameThresholdingAlgorithm{}} Algorithm}\label{alg:T_algorithm}
\end{algorithm}  

\section{Experimental evaluation} \label{s:results}
We will analyse the quality of our method through a set of metrics that allow us to determine  how well  the models 
capture the styles of performing a task observed in the expert executions (expressed as dependency graphs). More concretely, we will consider the following two quality measures:
fitness and simplicity. Given a set of dependency graphs $\mathcal{G}$ and a set of the models $\{M_i\}$, 
the fitness of each model $M_i$ is the number of dependency graphs of $\mathcal{G}$ with which $M_i$ overlaps. Note that if some dependency graphs overlap with a certain model is because they share a similar way  of solving the task. Hence, the fitness of a model is an indicator of how well the model captures such a way to execute the task. To compute the fitness of a solution (the set of inferred models), the models are applied in the order in that are generated. 
Lastly, the simplicity (or, alternatively, complexity) of a model is measured in terms of the number of edges and vertices that \modified{compose it.} Fitness and simplicity are well-known quality metrics of process discovery algorithms \citep{buijs2012role}. They are related to other state-of-the-art evaluation metrics in data science \citep{geng2006interestingness}: {\em Generality/Coverage} and {\em Conciseness} measures, respectively.  Additionally, the generality/coverage relation is employed in concept learning,  rule-learning and inductive logic programming to determine the examples that are matched by a hypothesis (model) \citep{raedt2010logic}. 



To experimentally evaluate our approach, we implemented a prototype using the programming language R\footnote{\url{https://www.r-project.org/}}. \textcolor{black}{The full code and data used for the experiments can be found in a github repository\footnote{\url{https://github.com/DNC87/MiningMultipleDependencyGraphs}}}.  We chose a challenging test-bed domain \textcolor{black}{as running example}: a typical suturing procedure in advanced surgeries like Laparoscopic Surgery. In laparoscopic exercises, one can find certain activities that can be considered as support manoeuvres.  Specifically, the surgical tasks that require many regrasping movements entail additional motions such as positioning or the tool reorientation which are not always really needed to perform the task. 
Additionally, the surgeon's idiosyncrasy and the diversity of surgery ``styles'' are another factors to be taken into account if we are interested in learning models in this domain. Our inductive method is able to account for all these factors during the learning of the models. \inserted{Finally, in order to test the versatility of the method, we have also applied it in a different domain, cooking brownies.}



\subsection{\textcolor{black}{Case Study 1: Surgical data}} \label{ss:Data}
There are not many public medical datasets specialised in laparoscopic exercises. Among the few available, possibly the most popular and complete is the JHU-ISI Gesture and Skill Assessment Working Set (JIGSAWS) dataset \citep{gao2014jhu}, which has been registered by using the surgery system robotics \textit{da Vinci}. Originally conceived for developing applications focused on the surgical motion analysis and the automatic skill assessment, this dataset has become a public benchmark for evaluating the performance of the state-of-the-art methods in surgical activity recognition \citep{ahmidi2017dataset}. 
Comparing with other human tasks, including surgery domain, the activities involved in suturing entail more complexity and diversity of movements than other training routines \citep{cao1996task}. 

\begin{figure}[t!]
\centering
  \includegraphics[width=1.0\columnwidth]{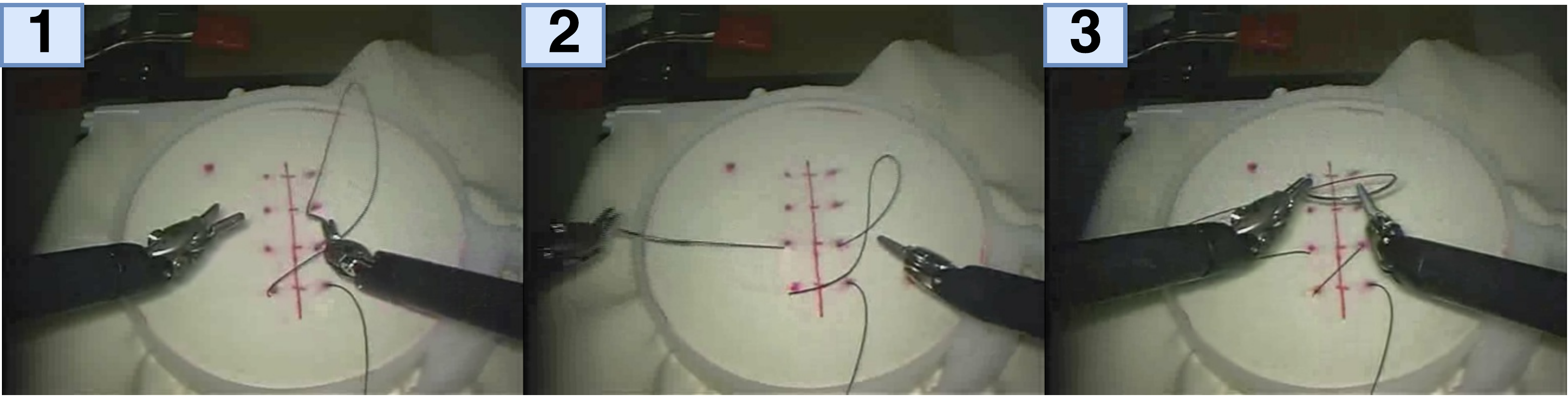}
  \caption{Example of a suturing cycle represented by frame steps. From left to right, frame \emph{(1)} shows the needle insertion in the input dot on the tissue, frame \emph{(2)} represents the needle grasping on the output dot from the other side of the tissue, and frame \emph{(3)} shows the needle transferring from left tool to the right tool. Captures obtained from JIGSAWS dataset \citep{gao2014jhu}.}~\label{fig:suturing_task_frames}
\end{figure}

We used the suturing executions provided by JIGSAWS dataset 
to learn the models from the different ways in which this task can be conducted. A simple suturing routine is divided into three phases. Firstly, the surgeon has to reach the needle and move it at the corresponding dot on the tissue. Then, the main phase of the exercise is performed: the suturing cycles. Figure \ref{fig:suturing_task_frames} shows this phase. For each cycle iteration (loop), the surgeon must push the needle against the next dot and take it out on the other side of the incision (Frame 1). The extraction of the needle must be carried out with the other hand (Frame 2). Then, the needle is transferred to the first hand again (Frame 3). Once the suturing cycles are completed, the last phase consists in laying the needle down at the finishing mark on the tissue. Concretely, for the experiments we chose a 4-throw suturing procedure, where the digit denotes the number of loops required to perform the suture.  


\begin{figure}[t!]
\centering
\begin{tabular}{ccc}
\includegraphics[width=0.6\columnwidth]{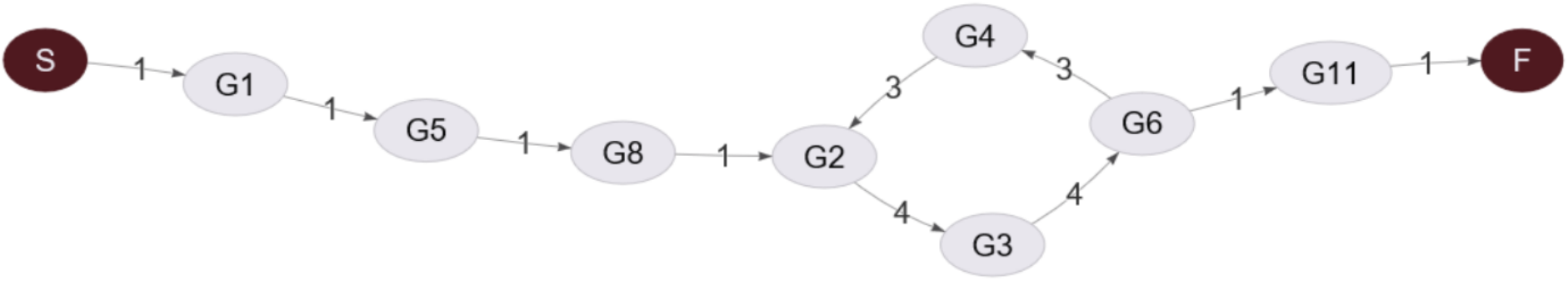}  \\
(a) Suturing example 1 \\[6pt]
\includegraphics[width=0.6\columnwidth]{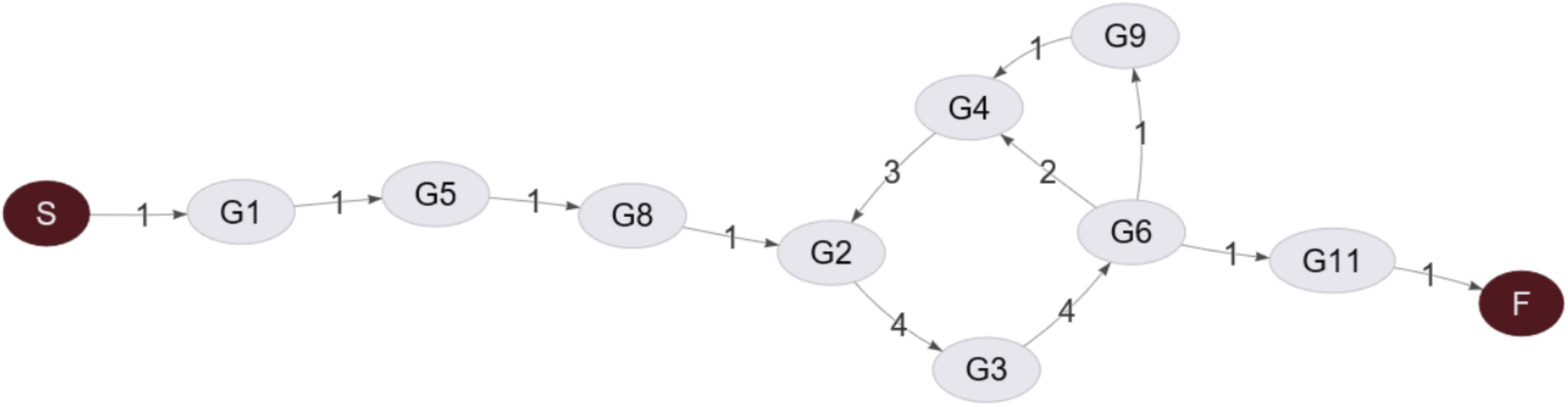}  \\
(b) Suturing example 2  \\[6pt]
\includegraphics[width=0.6\columnwidth]{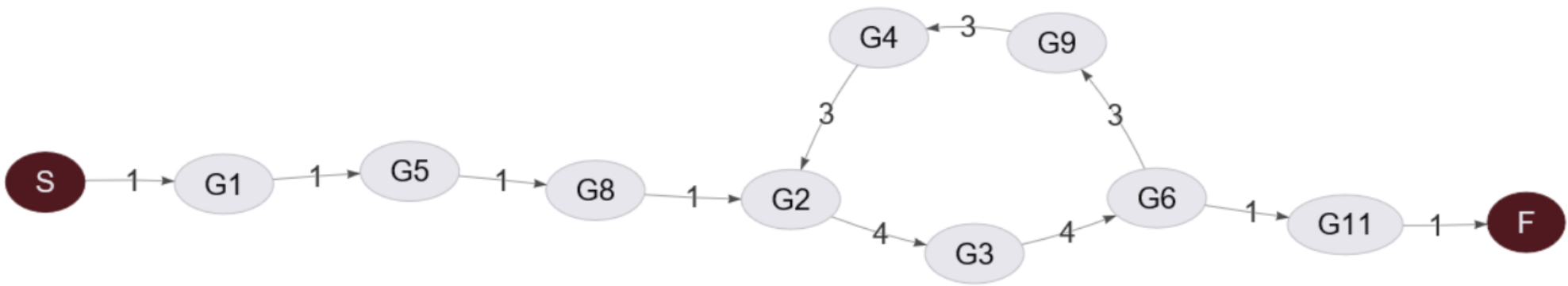}  \\
(c) Suturing example 3 \\[6pt]
\end{tabular}
\caption{Three trials from different surgeons expressed as dependency graphs. The differences reside in how each surgeon has performed the suturing loop. The labels of the vertices are provided in Table \ref{tbl:gesture_vocabulary}.}~\label{fig:trial_examples}
\end{figure}

In this context, an activity represents an atomic surgical gesture with a meaningful outcome and it is annotated following a specific activity language of this domain (Table \ref{tbl:gesture_vocabulary}). Hence, each activity transcription includes the name of the gesture, and the start and end frames in the video. Each performance contains from 15 to 20 activities per recording. It is necessary to mention that the activities of each task performance are annotated in chronological order of execution and there are not overlapping in time. Therefore, we consider each execution as a time-ordered sequence of labelled activities that a surgeon performed to completely fulfil the suturing task. Thus, we have preprocessed the transcription files to remove the time information and, then we have converted the activity sequences into dependency graphs.
\begin{table}[t!]
\centering
\small{
\begin{tabular}{ll}
ID & Description                             \\ \hline
G1            & Reaching for needle with right hand             \\
G2            & Positioning needle                              \\
G3            & Pushing needle through tissue                   \\
G4            & Transferring needle from left to right          \\
G5            & Moving to center with needle in grip            \\
G6            & Pulling suture with left hand                   \\
G7            & Pulling suture with right hand                  \\
G8            & Orienting needle                                \\
G9            & Using right hand to help tighten suture         \\
G10           & Loosening more suture                           \\
G11           & Dropping suture at end and moving to end points \\
G12           & Reaching for needle with left hand              \\
G13           & Making C loop around right hand                 \\
G14           & Reaching for suture with right hand            	\\
G15           & Pulling suture with both hands         			\\
\end{tabular}
}
\caption{Gesture vocabulary from JIGSAWS dataset \citep{gao2014jhu}. We consider those surgical gestures as activities throughout our  case study. 
}
\label{tbl:gesture_vocabulary}
\end{table} 
Figure \ref{fig:trial_examples} shows three trials 
extracted from the JIGSAWS dataset expressed as 
dependency graphs. 
As can be seen, the three graphs describe the suturing task composed by the needle preparation phase, the suturing loop, and the ending move where the needle is released. 
Notwithstanding the similarities, we can identify several differences between them. Thus, trials \emph{(a)} and \emph{(b)} seem very similar; however, there is an additional gesture ($G9$) in \emph{(b)} that is not in \emph{(a)}. This is the kind of support gesture we were referring to previously. Specifically, the surgeon has used the right tool to help tighten suture. 
Despite these manoeuvres are perfectly valid in the surgical domain, they may not be really necessary (as in trial \emph{(b)}), so they may be considered as noise. 
A different fact is when these supporting gestures are embedded inside the cycle as a routine activity as trial \emph{(c)} shows, which constitutes a way of performing the suturing task different of that exhibited by trials \emph{(a)} and \emph{(b)}. 

\begin{table}[]
\begin{subtable}{\textwidth}
\resizebox{1\textwidth}{!}{%
   \begin{tabular}{c|ccccccccccccccccccccccccccccccccccccccc}
Trial                                                                             & B001 & B002 & B003 & B004 & B005 & C001 & C002 & C003 & C004 & C005 & D001 & D002 & D003 & D004 & D005 & E001 & E002 & E003 & E004 & E005 \\ \hline
\begin{tabular}[c]{@{}c@{}}Global Rating Score\\ (GRS)\end{tabular} & 13   & 17   & 14   & 10   & 12   & 26   & 20   & 26   & 30   & 17   & 14   & 18   & 14   & 8    & 15   & 17   & 20   & 19   & 19   & 19   \\ \hline
\begin{tabular}[c]{@{}c@{}}Surgeon expertise\\ (in robotic surgery)\end{tabular}               & N    & N    & N    & N    & N    & I    & I    & I    & I    & I    & E    & E    & E    & E    & E    & E    & E    & E    & E    & E    \\ \hline
\begin{tabular}[c]{@{}c@{}}Scoring Quartile\end{tabular}                        & Q4   & Q3   & Q4   & Q4   & Q4   & Q1   & Q2   & Q1   & Q1   & Q3   & Q4   & Q3   & Q4   & Q4   & Q3   & Q3   & Q2   & Q2   & Q2   & Q2   \\ \hline
\end{tabular}}
\end{subtable}
\newline
\vspace*{0.3 cm}
\newline
\begin{subtable}{\textwidth}
\resizebox{1\textwidth}{!}{%
\begin{tabular}{c|ccccccccccccccccccccccccccccccccccccccc}
Trial                                                                             & F001 & F002 & F003 & F004 & F005 & G001 & G002 & G003 & G004 & G005 & H001 & H003 & H004 & H005 & I001 & I002 & I003 & I004 & I005 \\ \hline
\begin{tabular}[c]{@{}c@{}}Global Rating Score\\ (GRS)\end{tabular} & 24   & 26   & 29   & 24   & 29   & 13   & 18   & 13   & 21   & 23   & 14   & 25   & 19   & 21   & 17   & 23   & 17   & 23   & 19   \\ \hline
\begin{tabular}[c]{@{}c@{}}Surgeon expertise \\ (in robotic surgery)\end{tabular}               & I    & I    & I    & I    & I    & N    & N    & N    & N    & N    & N    & N    & N    & N    & N    & N    & N    & N    & N    \\ \hline
\begin{tabular}[c]{@{}c@{}}Scoring Quartile\end{tabular}                        & Q1   & Q1   & Q1   & Q1   & Q1   & Q4   & Q3   & Q4   & Q2   & Q1   & Q4   & Q1   & Q2   & Q2   & Q3   & Q2   & Q3   & Q2   & Q3   \\ \hline
\end{tabular}}%
\end{subtable}
\caption{Trial information for suturing task in JIGSAWS dataset \cite{gao2014jhu}. The rows represent: the surgeon's experience in robotic surgery (measured in hours): E-expert (>100hrs), I-intermediate (10--100hrs), N-novice (<10hrs); the GRS score reported by the expert reviewers, and the GRS score in quartile partitions. The columns correspond to the trials, where the letter identifies the surgeon who performed the trial and the number denotes the trial order.}\label{tbl:trial_scores}
\end{table}

Finally, Table \ref{tbl:trial_scores} 
shows the information about the executions or ``trials'' (term used in the dataset to identify a task execution) provided by the dataset. To improve the readability of this table, the columns represent trials and the rows are their description. A trial is identified by a code, formed by the surgeon identifier (a letter from B to I) and the order of her trial (e.g. H003 means that surgeon H has recorded her trial number 003). As can be seen, eight surgeons performed this task 5 times which makes a total of 40 trials. The rows in the table show the metadata that JIGSAWS provided for describing each trial: the global rating score (GRS\footnote{More information about this measure in \citep{gao2014jhu}}) which measures the technical skill over the entire trial; the surgeon's level of expertise in robotic surgery  measured in hours of practice (i.e., E-expert (>100hrs), I-intermediate (10-100hrs), N-novice (<10hrs)), and  the scoring quartile that we added by stratifying the scores into quartiles. All of this information will be used to analyse the performance of our method for different trials selection criteria.


\subsubsection{Experimental Setup} \label{ss:exp_setup}

The input dependency graphs have been divided in two groups: the training and the test sets.  Given that we have few examples, we  prepared three exploratory scenarios applying different criteria to select the training graphs and  analysed the impact of the training data selection in the learning process. For the first scenario (Experiment 1), we split the data according the surgeon's expertise level 
using the trials where surgeon's tag is ``E'' (Expert) as training graphs. For the second and third scenarios (Experiments 2 and 3, respectively), we used the GRS score as a criterion for splitting the data.  
Specifically, as training data we used the dependency graphs belonging to the $Q1$ quartile for Experiment 2, and the dependency graphs belonging to the $Q1$ and $Q2$ quartiles for Experiment 3. Experiments 2 and 3 were carried out to analyse the impact of the size of the training data on the inferred models. 
Thereby, the number of training dependency graphs used for each experiment was $n = 10$ (Experiment 1), $n = 9$ (Experiment 2) and $n = 16$ (Experiment 3). In each of these scenarios, those dependency graphs not used for training were used as test set. 

In what follows, we denote the 
models using Roman alphabet upper-case letters in ascending order (e.g., $A_i$, $B_i$, $C_i$, \ldots{}),  according to the order in that they are  generated, and where the subindex $i$ identifies the experiment ($i\in\{1, 2, 3\}$). 


\subsubsection{Experimental results} \label{ss:exp_results}

\begin{figure}[ht!]
\centering
  \includegraphics[width=0.8\columnwidth]{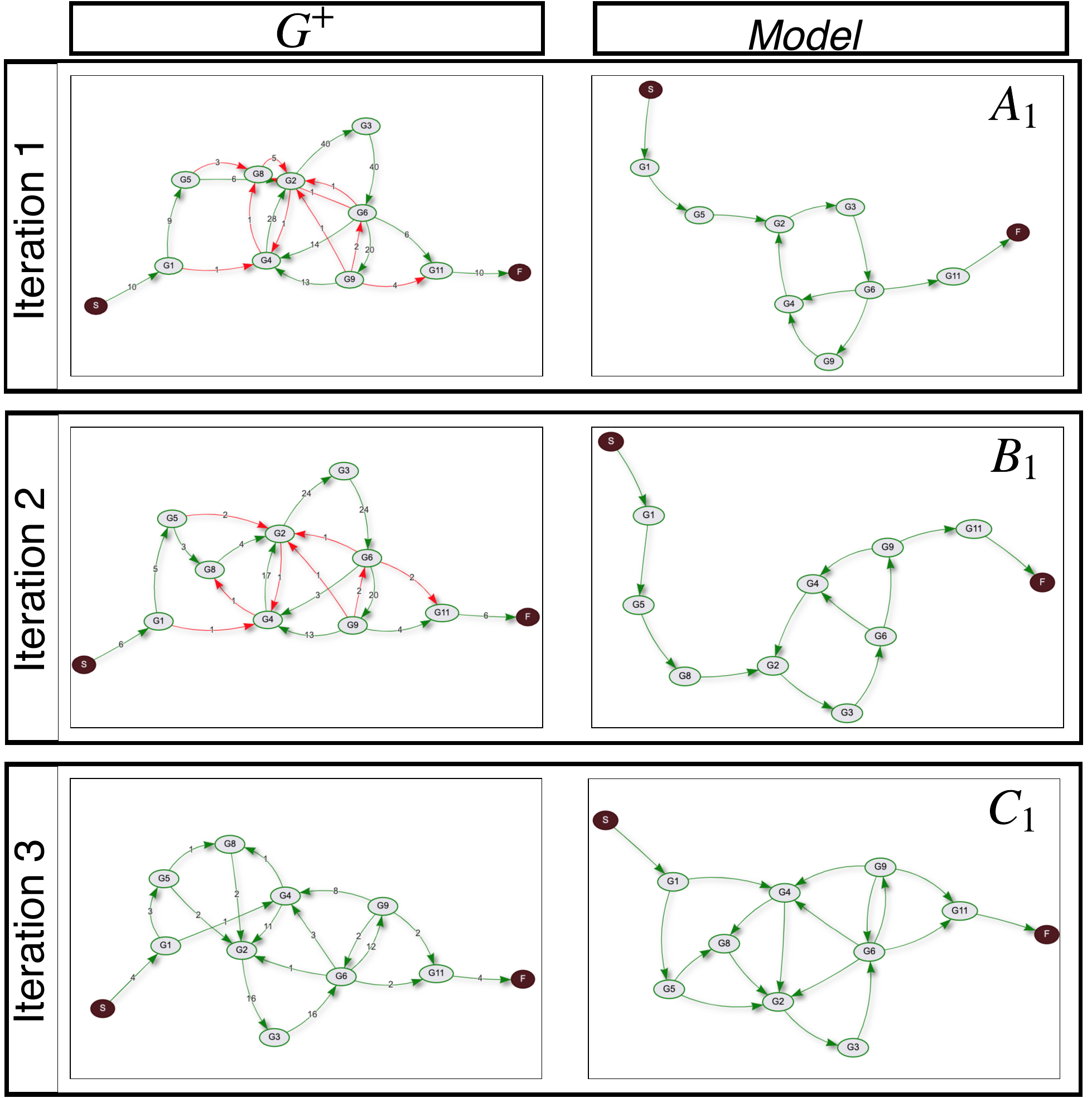}
  \caption{Models 
  inferred from the Expert's dependency graphs  (Experiment 1). Rows contain the corresponding $\AggregatedGraph$ and model generated at each iteration of the {\scshape\ShortNameMiningAlgorithm{}} algorithm. We have coloured in green those edges in $\AggregatedGraph$ that remain in the model. By contrast, the edges coloured in red are those that are removed by the algorithm when refining $\AggregatedGraph$.}~\label{fig:expert_workflows}
\end{figure}

Before reporting the performance obtained by our method in the experimental evaluation, we show in Figure \ref{fig:expert_workflows} the models 
learnt in  Experiment 1.

As can be seen, we can clearly identify the phases of a suturing exercise in models $A_1$ and $B_1$: the needle reaching, the suture cycle and the needle releasing at the finishing mark. Although both models are very similar, we detect two important differences: $B_1$ incorporates the  activity $G_8$ (i.e., reorienting the needle) before the suturing cycle and the activity $G_9$ (i.e., using the right hand to tighten the suture) just before releasing the needle. In contrast, in the third iteration, no edges were removed from $\AggregatedGraph$ to generate model $C_1$. As a consequence $C_1$ is a large and complex model that captures more specific ways of performing the suturing exercise.



\begin{table*}[ht!]
\centering
\resizebox{0.9\textwidth}{!}{%
 \def\arraystretch{.7}
    \begin{minipage}[t]{0.33\linewidth}
      \caption*{Experiment 1 \\ (Experts)}
      \centering
        \begin{tabular}[t]{cc}
        ID   & Model \\ \hline
        E001 & $A_1$        \\
        E002 & $A_1$        \\
        E003 & $A_1$       \\
        E005 & $A_1$       \\
        D003 & $B_1$       \\
        D005 & $B_1$        \\
        D001 & $C_1$        \\
        D002 & $C_1$        \\
        D004 & $C_1$        \\
        E004 & $C_1$        \\ \hline
        \multirow{3}{*}{\begin{tabular}[c]{@{}c@{}}Summary\\ (n=10)\end{tabular}} 
        & A: 4 \\
        & B: 2 \\
        & C: 4
        \end{tabular}
    \end{minipage}%
    
    \begin{minipage}[t]{0.33\linewidth}
      \centering
        \caption*{Experiment 2 \\ (GRS (Q1))}
        \begin{tabular}[t]{cc}
        ID   & Model \\ \hline
        C001 & $A_2$        \\
        C003 & $A_2$        \\
        C004 & $A_2$        \\
        F005 & $A_2$        \\
        H003 & $A_2$        \\
        F003 & $B_2$        \\
        F004 & $B_2$        \\
        F001 & $C_2$        \\
        F002 & $C_2$        \\ \hline
        \multirow{3}{*}{\begin{tabular}[c]{@{}c@{}}Summary\\ (n=9)\end{tabular}} 
        & A: 5 \\
        & B: 2 \\
        & C: 2
        \end{tabular}
    \end{minipage} 
    
    \begin{minipage}[t]{0.33\linewidth}
      \centering
        \caption*{Experiment 3 \\ (GRS(Q1+Q2))}
        \begin{tabular}[t]{cc}
        ID   & Model \\ \hline
        C001 & $A_3$          \\
        C003 & $A_3$        \\
        C004 & $A_3$         \\
        F005 & $A_3$        \\
        H003 & $A_3$        \\
        I002 & $A_3$         \\
        I004 & $A_3$         \\
        C002 & $B_3$         \\
        F003 & $B_3$         \\
        F004 & $B_3$         \\
        G004 & $B_3$         \\
        G005 & $B_3$         \\
        H005 & $B_3$         \\
        E002 & $C_3$         \\
        F001 & $C_3$         \\
        F002 & $C_3$         \\\hline
        \multirow{3}{*}{\begin{tabular}[c]{@{}c@{}}Summary\\ (n=16)\end{tabular}} 
        & A: 7 \\
        & B: 6 \\
        & C: 3
        \end{tabular}
    \end{minipage}} 
    \caption{For each experiment, the table contains the training dependency graphs (column ID) and the model which overlaps with them (column category). In Experiment 1, the training data corresponds to the trials performed by the experts (a total of $n=10$ instances). In Experiments 2 and 3 the training data were selected according to their GRS score quartiles: the first quartile ($Q1$) in Experiment 2 (a total of $n=9$ instances), and the first and the second quartiles ($Q1$ + $Q2$) in Experiment 3 ($n=16$ instances in total). At the bottom of each table it is shown the fitness summary per experiment.} 
\label{tbl:training_results}
\end{table*}

Table \ref{tbl:training_results} shows the results of the learning process  and the fitness analysis with respect to the training data for the three experiments. 
Firstly, we observe that, by chance, 3 models were inferred in each experiment. Regarding the effect of the size of the training data set, although Experiment 3 used a number of training graphs greater than the other two experiments, no more models were obtained. Comparing Experiments 2 and 3 (which were conceived using a selection criteria of the training data based on the score quartiles) we can conclude that the size of the training data set per se does not seem to have a direct influence on the size of the solution (number of models), but the similarity among the training data. It can be assumed that this similarity depends to a large extent on the suturing skills of the surgeon which is measured by the GRS value. Thus, the dependency graphs belonging to quartiles $Q1$ and $Q2$ should  be similar in the sense that they not show any new popular way of performing the task apart from those expressed by the examples used in Experiment 2. 

If we compare Experiment 1 and Experiment 2, we observe similar results between them, regarding the fitness of the models. Perhaps the most remarkable fact is that the fitness of model $C_1$, the last model generated from the experts executions (Experiment 1) is higher than the fitness of its  counterpart  model $C_2$  from Experiment 2. This means that the training data is more uniform in Experiment 2 than in Experiment 1 (that is, there are more training graphs in Experiment 2 that follow the same styles of solving the suturing task which are captured by models $A_2$ and $B_2$).  

\begin{figure}[ht!]
\centering
\begin{subfigure}[b]{0.8\textwidth}
   \includegraphics[width=1\linewidth]{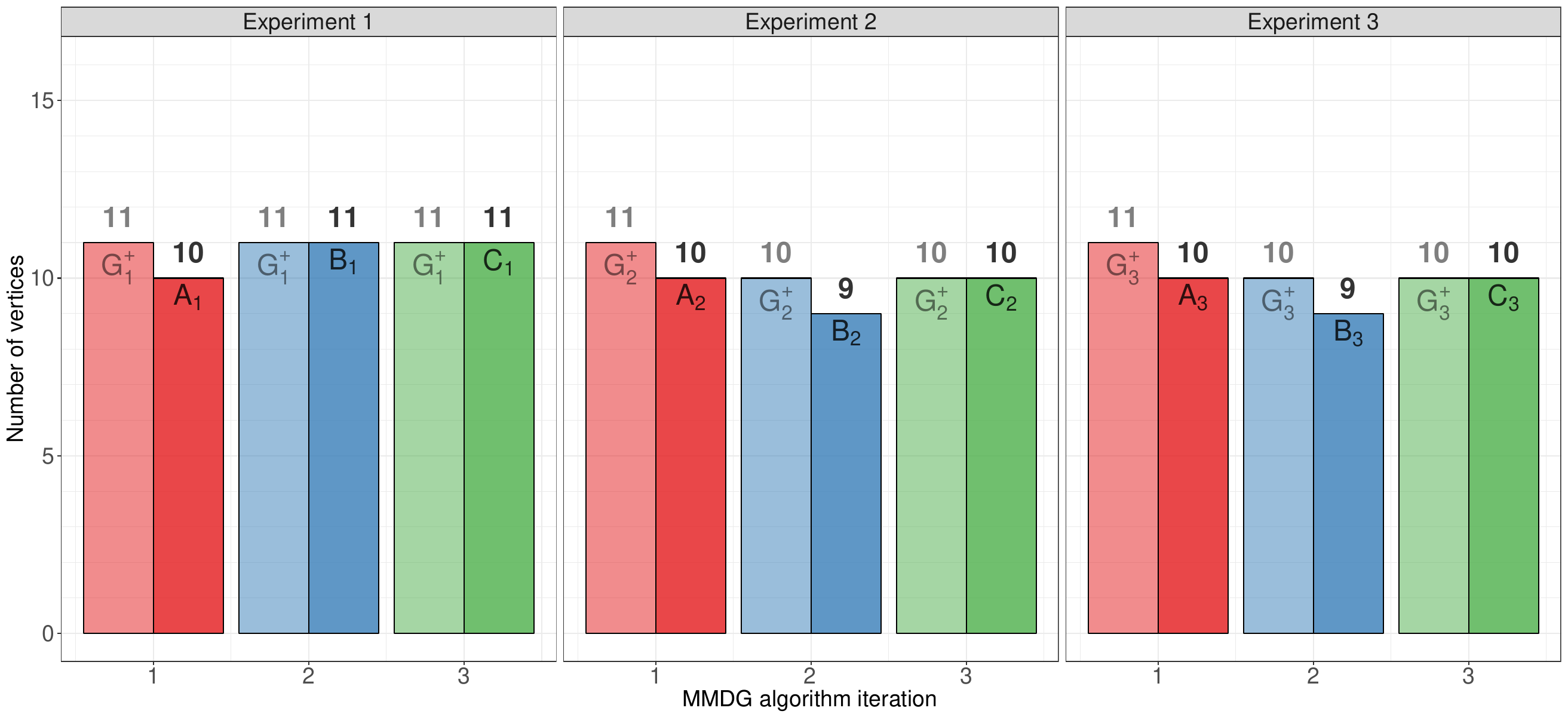}
   \caption{\scriptsize Number of vertices in $G^+_i$ and models $A_i$, $B_i$ and $C_i$ for each experiment.}
   \label{fig:vertices_edges_comparison_vertices} 
\end{subfigure}

\begin{subfigure}[b]{0.8\textwidth}
   \includegraphics[width=1\linewidth]{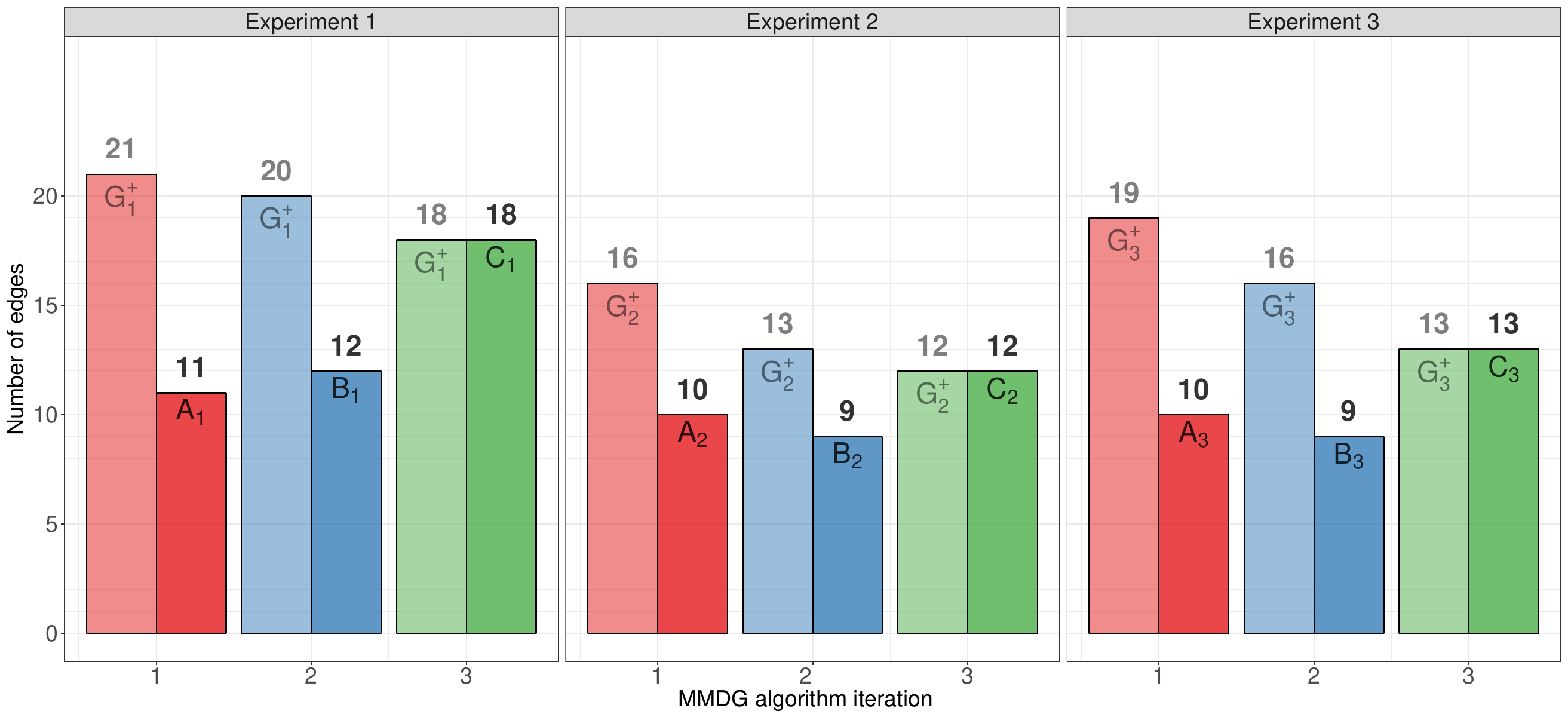}
   \caption{\scriptsize Number of edges in $G^+_i$ and models $A_i$, $B_i$ and $C_i$ for each experiment.}
   \label{fig:vertices_edges_comparison_edges}
\end{subfigure}
\caption{Number of vertices (top) and edges (bottom) of the aggregate graphs and the models generated during their refinement for  Experiments 1, 2 and 3. The algorithm iterations have been distinguished using colours (iteration 1 - red, iteration 2 - blue and iteration 3 - green) and the results have been grouped by experiment. 
}~\label{fig:vertices_edges_comparison}
\end{figure}

The analysis of the simplicity of the models 
is depicted in Figure \ref{fig:vertices_edges_comparison}. Specifically, we show the number of vertices (top) and edges (bottom) of each induced model. The smaller the number of vertices and edges, the simpler the model is. In all the experiments, we observe that, as expected, some vertices (i.e., activities) and edges (i.e., transitions between activities) were removed during the process of refining the aggregated graphs $\AggregatedGraph_i$  to produce the models ($A_i$/$B_i$/$C_i$).

Concretely, regarding the results with respect to the vertices, it is observed a slight reduction in their number. Thus, in all the models generated in iterations 1 and 2, one vertex was filtered 
except in $B_1$. This means that one unnecessary activity to perform the task was discarded, with the subsequent positive effect on model compactness. On the other hand, the results regarding the edges show a noticeable reduction of them in models 
$A_i$ and $B_i$ (comparing them with their respective aggregated graphs). 
Specifically, the number of edges of $A_1$ decreases by 47.62\% with respect to its aggregated graph. 
Similar decreases are observed in other models, such as $A_3$ (47.37\%), $B_3$ (43.75\%), $A_2$ (37.50\%), $B_1$ (40.00\%) and $B_2$ (30.77\%). All this has a positive effect on the compactness of the inferred models. 

Additionally, if we compare the three experiments, we observe that using the trials executed by the experts (Experiment 1), the model $\AggregatedGraph_1$ is the aggregated graph with more edges which means that this is the graph that contains more 
non-essential transitions between activities. These dependency graphs are depicted in Figure \ref{fig:expert_workflows}. Analogously, the models 
learned in Experiment 1 were those that had the most activities and transitions, if we compare them with the respective models 
obtained in the other two experiments. This is further supported by the difference in the number of edges between  model $C_1$ and  models $C_2$ and $C_3$. The last models generated by the \ShortNameMiningAlgorithm{} algorithm usually gather the non-essential transitions  filtered in previous iterations. In this case, we observe that more edges were gathered in the case of the experts. Similar conclusions are derived from the analysis of the number of vertices in Experiment 1. 
Based on these results, we can conclude that the trials recorded by the experts are more noisy (they contain many actions that are  non-essential for performing the task). However, our approach is able to infer models that filter that noise in all the experiments.

We also analysed how the size of the training data set affects the simplicity of the models. 
As mentioned above, to this end we compared Experiments 2 and 3. 
Thus, in the two first iterations of the \ShortNameMiningAlgorithm{} algorithm the aggregated graphs had more edges in Experiment 3 than in Experiment 2. Nevertheless, the generated models 
had the same number of edges and vertices in both experiments. That means that the $Q2$ training graphs only contribute with transitions between activities that indeed are non-essential for the task since they were not included in the models. 

Finally, to study the generalisation capacity of our models we calculated their fitness over a set of unseen dependency graphs (the test set). Table \ref{tbl:test_results} shows the fitness  results for the test data. 

\begin{table}[ht!]
\centering
\begin{tabular}{c|c|cc|cc|cc|cc}
Experiment & n & $A_i$ & \% & $B_i$ & \% & $C_i$ & \% & $\text{-}_i$ & \% \\ \hline
1 & 29 & 2 & 6,90\% & 0 & 0,00\% & 16 & 55,17\% & 11 & 37,93\% \\
2 & 30 & 11 & 36,67\% & 6 & 20,00\% & 1 & 3,33\% & 12 & 40,00\% \\
3 & 23 & 9 & 39,13\% & 2 & 8,70\% & 5 & 21,74\% & 7 & 30,43\%
\end{tabular}
\caption{Fitness of the inferred models over the test data. The columns 1 and 2 show the experiment identifiers and the size of the test set, respectively. The next three column blocks correspond to the absolute (denoted as $A_i$, $B_i$ and $C_i$) and relative (denoted as $\%$) fitness value per model. The last block of columns ($\text{-}_i$ and $\%$) shows the absolute and relative number of test graphs that do not overlap with any model.}
\label{tbl:test_results}
\end{table}

In general terms, we observe that the global fitness rate (adding the fitness of all the models 
induced in an experiment) is in the range between 60\% and 70\% for all the experiments. If we observe the fitness achieved by the models 
induced from the experts (Experiment 1), the fitness rate of $A_1$ and $B_1$ is very low. Despite that around 62\% of test graphs overlap with the models in this experiment, only 6.90\% of the test data overlap with a different model to $C_1$. The high fitness rate of $C_1$ may be motivated by the fact that the last generated model usually gather more edges and activities, so it can overlap with many more examples. However, $C_1$ is the poorest model in terms of simplicity. By contrast, Experiment 2 shows a greater fitness rate in models $A_2$ and $B_2$, but not $C_2$, compared to Experiment 1. Although  Experiment 2 used a smaller training set than Experiment 1, we do not observe any effect on the rate of test data that do not overlap with any model. Specifically, the rise is only about 2\% over the rate reached in the Experiment 1 ($\text{-}_1$ = 37.93\% and $\text{-}_2$ = 40\%).

On the other hand, the shift of the fitness rate towards models $A$ and $B$ in Experiment 2 is an interesting outcome. 
Models $A_2$ and $B_2$ have been more suitable to generalise the task, since they overlap with a noticeable portion of the examples not seen during the model training ($A_2$ = 36.67\% and $B_2$ = 20\%). The fitness rates reached by models $A_2$ and $B_2$ are significantly larger than those obtained by models $A_1$ and $B_1$ (Experiment 1). However, the fitness results of Experiment 3 show a different picture. 
Firstly, we can see how the number of test graphs that do not overlap with any model decreases from 40 \% (obtained in Experiment 2) to 30.43 \%, whereas most of the test graphs overlap with models 
$A_3$ and $C_3$. Although the fitness rate of $A_3$ is higher than that of $A_2$, the difference is only of 2.46\%. Conversely, the increment in the fitness of $C_3$ with respect to $C_2$ is of 18.41\%. All this corroborates the fact observed with the training data that by adding the $Q_2$ dependency graphs to the training set, 
only contributes with graphs with more non-needed activities/transitions (that is, noise), which in turn help model $C_3$ to  fitness more graphs. 
Nonetheless, even in the best case of  fitness (obtained by Experiment 3), almost 30\% of the test graphs show way of performing the suturing task that are not gathered by the models.

\subsection{\inserted{Case Study 2: brownie cooking}} 
\inserted{We have applied the MMDG algorithm to another challenging problem in a different domain: cooking. This domain is especially challenging for different reasons: the examples have high variability among them and they do not contain extra information about the quality of the execution. The brownie cooking dataset \citep{spriggs2009temporal} consists of 16 executions from different users and, at most, 11 different activities per execution. The scarcity of data comes not only from the number but also the length of the examples. Table \ref{tbl:brownie_vocabulary} shows the vocabulary of the brownie cooking dataset.}

\begin{table}[t!]
\centering
\small{
\begin{tabular}{ll}
ID & Description                             \\ \hline
A\_4            & crack-egg             \\
A\_12            & pour-big-bowl-into-baking-pan                              \\
A\_13            & pour-brownie-bag-into-big-bowl                   \\
A\_14            & pour-oil-into-big-bowl          \\
A\_16            & pour-water-into-big-bowl            \\
A\_26            & spray-pam                   \\
A\_27            & stir-big-bowl                  \\
A\_28            & stir-egg                                \\
A\_29            & switch-on-the-oven         \\
A\_31           & take-big-bowl                           \\
A\_33           & take-egg \\
\end{tabular}
}
\caption{\inserted{Vocabulary of the brownie cooking dataset.} 
}
\label{tbl:brownie_vocabulary}
\end{table}

\begin{figure}[ht!]
\centering
  \includegraphics[width=0.8\columnwidth]{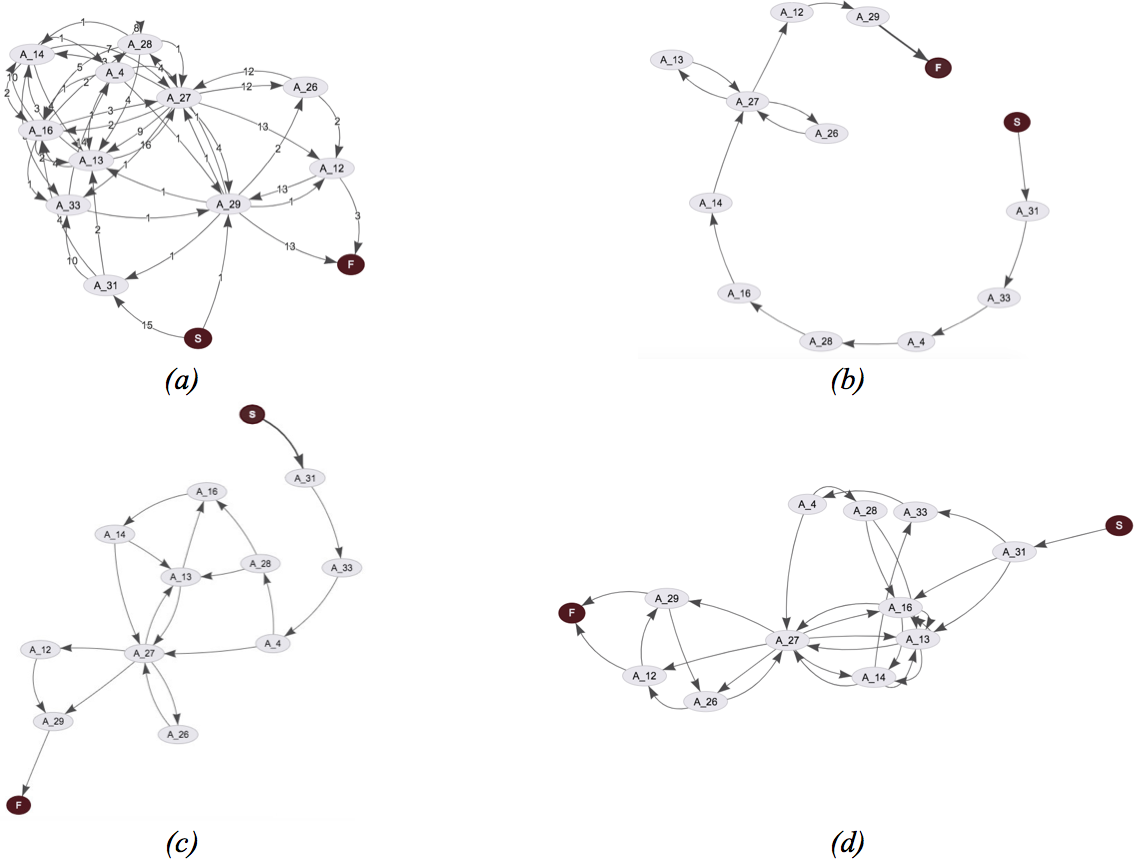}
  \caption{\inserted{Brownie example. (a) aggregated graph with all examples, (b) model generated on iteration 1 and threshold 7, (c) model generated on iteration 2 and threshold 4, and (d) model generated on iteration 3 and threshold 2.}}~\label{fig:Brownie_example}
\end{figure}

\inserted{Our algorithm  obtains four models. Figure \ref{fig:Brownie_example} shows the aggregated graph and the first three models. 
The fourth model (not shown) includes all the variability not contemplated by the three previous models, as has been described previously.
With only these 16 examples, MMDG can extract three ways of cooking the brownie that allows for the understanding and supervising new executions of the task. As can be seen in Figure \ref{fig:Brownie_example}, as the threshold decreases, the variability of the model increases, giving rise to increasingly complex graphs. 
The model obtained in the first iteration and corresponding to a threshold of 7 (Figure \ref{fig:Brownie_example}.b) practically represents a sequence of activities and would correspond to a strict monitoring of the recipe. In the second model obtained with a threshold of 4 (Figure \ref{fig:Brownie_example}.c), once the egg has been introduced into the bowl (which coincides with the first steps of the simplest model), the model has presented some variations with respect to strict monitoring of the recipe. Finally, the third model obtained with a threshold of 2 (Figure 9.d) is the most complex and presents execution variability from the very first moment.}

\begin{figure}[ht!]
\centering
  \includegraphics[width=0.8\columnwidth]{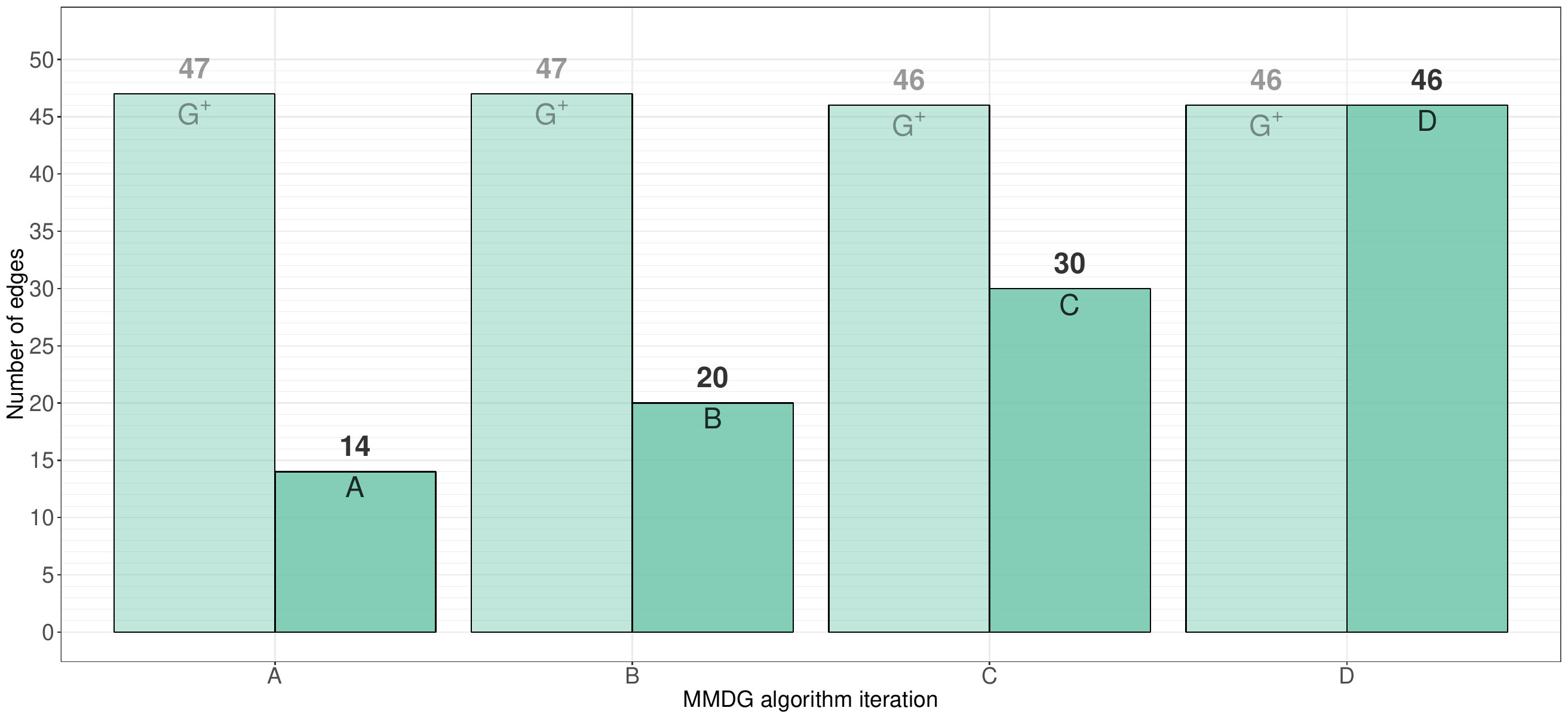}
  \caption{\insertedd{Brownie example. Comparison of the number of edges between the aggregate graph (pale green) and the model (dark green) that was obtained as a result for each iteration of our algorithm.}} 
  \label{fig:Brownie_simplicity}
\end{figure}

\insertedd{The analysis of the simplicity of the models of the brownie example is depicted in Figure \ref{fig:Brownie_simplicity}. In this example, the induced models show a quite pronounced reduction of the number of edges in the first (marked as A) and second (marked as B) induced models. These show a reduction of 70\% (14 out of 47) and 67\% (20 out of 47) respectively. In the case of the third model (marked as C) the reduction is 45\% (30 out of 46). The fourth model, as expected, has no edge reduction. Regarding the results with respect to the vertices, there is no reduction in any of the obtained models. 
This is because in these examples all the activities are present in the 16 examples and only the transition between executions (the edges) are different.}     

\insertedd{With respect to fit analysis, this dataset contains only sixteen examples, thus it is not convenient to separate between training and testing examples. Therefore, the results  correspond to the whole dataset. The algorithm has been able to identify 3 different ways (with increasing level of complexity/variability) having only 16 examples.  
The results indicate that only one of the examples fits the first model, three examples fit  the second model and  three fit the third model. The fourth and most complex model, which collects all the variability of the executions, includes the rest of examples (9 out of 16). 
}

\section{Discussion} \label{s:discussion}


\modified{Given the 
existence of process mining tools \citep{van2017rapidprom}, a first question is how this compare to them.} 
In the case of learning a workflow model, the main goal of process mining is to model the process that underlies in the event logs, but not to extract a set of models with the different forms of the task and the essential activities and transitions in them. 
In principle, they do not assume that \modified{there may} be noise, non-essential or missing activities in the process logs. Then, the whole processes are modelled. \modified{This way of doing process modelling often leads} to spaghetti-like models \citep{gunther2007fuzzy}. Therefore, the commercial tools like Fluxicon DISCO\footnote{https://fluxicon.com/disco/} includes options that \modified{allow for the simplification of} the obtained model based on the ``popularity'' of the edges. However, the final result is only one process model and it is not able to extract all the ways of performing a task. In addition, tools like DISCO do not check that the simple model obtained is supported by one example at least.

\begin{figure}[ht!]
\centering
  \includegraphics[width=0.8\columnwidth]{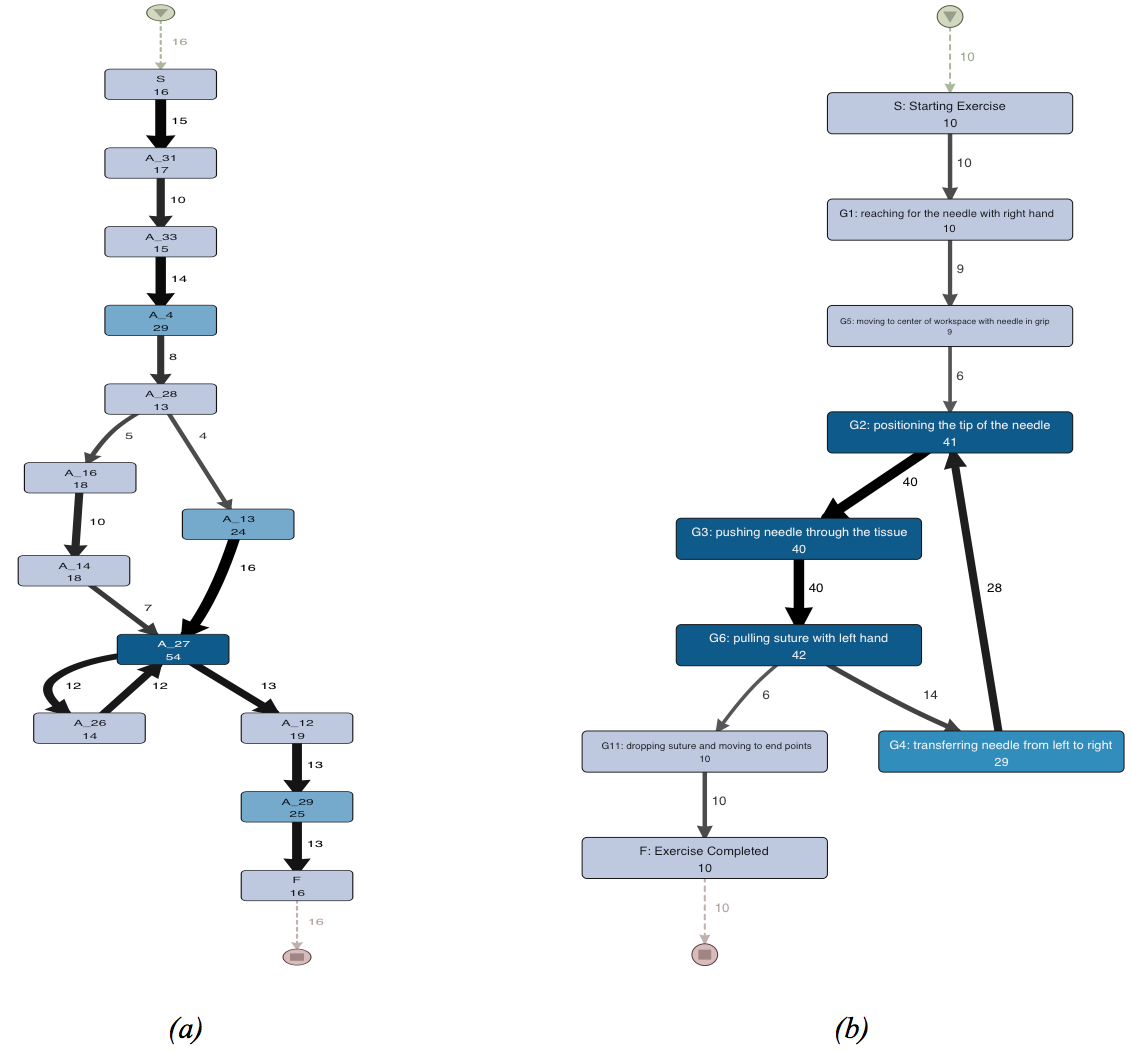}
  \caption{\inserted{Models obtained by DISCO. (a) Brownie executions model, (b) Suture executions model.}}~\label{fig:DISCO}
\end{figure}

\inserted{We can see some of these issues after applying DISCO to the data for the surgical and cooking domains. Figure \ref{fig:DISCO} shows that DISCO only obtains one model from the aggregated graph and it is obtained by collapsing edges and nodes depending on the chosen threshold (manually determined). According to the algorithm description, the DISCO model obtained from the brownie examples has two disjoint paths in the middle of the model. In one path the user must put water (A\_16) and oil (A\_14) in the bowl but not the brownie bag (A\_13). In the other path the user must put the brownie bag but not the water and the oil. These execution options, which are clearly wrong, were not possible with the MMDG algorithm because we check that the execution can be completed correctly following the model. Besides, we can extract not only one but all the model executions that have a minimum number of examples.}

\inserted{Figure \ref{fig:DISCO} also shows the model for the suture example (see Section \ref{ss:Data}) obtained using DISCO.  In this case, the result is very similar to the model $A_1$ inferred from iteration 1 of MMDG applied on expert executions. However, some differences can be observed. First of all, it obtains only one model, discarding, in this way, the differences among expert executions. Second, the activity G9 does not appear, although it is executed by most of the experts in some iterations to assist in the tightening of the thread. Finally, the threshold to obtain the model must be set manually. However, this parameter is automatically set in MMDG.}

\begin{figure}[ht!]
\centering
  \includegraphics[width=0.8\columnwidth]{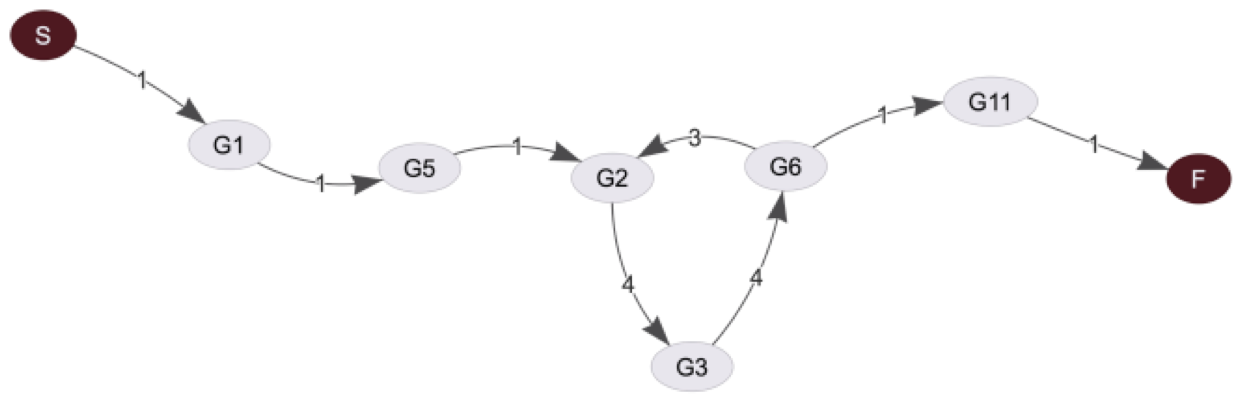}
  \caption{\inserted{Fresh example of execution of the suture exercise.}}~\label{fig:New_execution}
\end{figure}

\begin{figure}[ht!]
\centering
  \includegraphics[width=0.8\columnwidth]{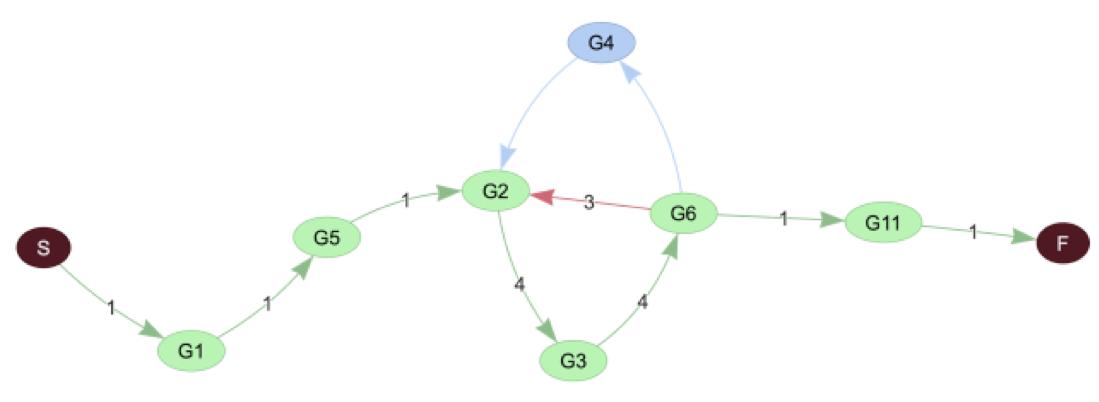}
  \caption{\inserted{Result of replaying the new exercise (Figure \ref{fig:New_execution}) in the first model obtained from the suture examples (see Section \ref{ss:exp_results}). The wrong transitions are marked in red (from G6 to G2), while the missing ones are marked in blue (from G6 to G4 and from G6 to G2). }}~\label{fig:Supervision}
\end{figure}

Different process mining algorithms have been also \modified{analysed, such as the Alpha-algorithm \citep{alves2004process} or fuzzy models \citep{gunther2007fuzzy}}. But unlike the model that can be obtained by these other methods, our ability to extract and manage several representations for the same task could be particularly beneficial for supervision. In this way, we can provide more accurate supervision to the user if we can identify their particular way of carrying out the task. \inserted{To supervise a fresh example of task execution using our models, it is only necessary to replay \citep{van2016process} it in any of the models obtained. 
Figure \ref{fig:New_execution} shows a dependency graph of a fresh execution of the suture exercise. If we replay this execution in the first model obtained for the suture task, we will obtain the results shown in Figure \ref{fig:Supervision}. As this figure shows, by replaying the example in the model, you can detect erroneous transitions (marked in red) between the activities and those that are missing (marked in blue) and that should be carried out in order to be successful in the execution of the exercise. } 

\modified{Our major aim with this paper was to} build a system capable to learn the task models in contexts with very few demonstrations and having only positive examples. This is why it is required that these examples be as meaningful and rich as possible. This is supported by the results of experiments 1 and 2 set out in Section \ref{ss:exp_setup} and analysed in Section \ref{ss:exp_results}. As these experiments demonstrates, the better the GRS score of the exercises used during learning, the better the models obtained that is shown by the quality measures analysed. Thus, \modified{we have to be very careful}  in selecting the correct examples for the training set. \modified{The third experiment set out in Section \ref{ss:exp_setup} answered the question of how many examples are needed for learning the process models. In this case, we applied} the 
algorithm to one training set that consists of the examples of the first and the second quartile based on the GRS score. What we have observed is that the resulting models do not improve the quality measures analysed in experiment 2 (that only uses examples of the first quartile). Moreover, the obtained models in experiment 3 are very similar to the ones obtained in experiment 2.

Finally, one may wonder if the use of dependency graphs could introduce a bias to the solution that limits its expressiveness. As we have mentioned, dependency graphs are an easy representation of process models that facilitate their simplification. Besides, there \modified{are tools that ease} the further transformation of those models into a high-level representation such as Petri nets or BPMN (i.e., Business Process Model and Notation), or into a logic \modified{program} with interesting features such as reasoning about scenarios as Event Calculus does \citep{cicekli2000formalizing} but that require quite more complex simplification/abstraction tools. 

\modified{To close the discussion, we clarify} the limitations of our approximation. First of all, we lose important information regarding the task during our mining method (e.g., the number of times the suturing cycle must be done) when we perform the aggregation of the examples. However, this meaningful information could be recovered by replaying the covered demonstrations on the corresponding obtained model. On the other hand, the semantics of an activity that is repeated several times in a sequence can vary throughout a task. This is part of representational bias that we assume by using dependency graphs. This type of representation is potentially limited to represent concurrency, duplicate actions, hierarchical activities, or model OR-splits/joins in a mined model. For instance, \citet{yang2017medical} proposed a method to split duplicate activities in the final model in order to gain expressiveness. This approximation could be introduced in future versions of our algorithm. Finally, we have observed that many examples did not overlap with any model in the experiments (30\% and 40\% of examples). We think our criterion to determine when an example conforms to a model may be very strict.
Therefore, other less restrictive conformance checking will be analysed in the future to reduce these cases.

\section{Conclusions and Future Work} \label{s:conclusions}

In this paper, we have presented a new approach \modified{to learn the different strategies for} performing a task based on a few 
examples. 
Our proposal starts from an event log with the set of executions of a task in the form of sequences of activities. After converting them into dependency graphs, we apply our novel inductive method for learning models from these dependency graphs.  
As a result, we can generate the multiple models 
with the essence of the task, eliminating the noise (i.e., unnecessary transitions between activities). \modified{We have applied our approach to two challenging task: suture in minimally invasive surgery and cooking a brownie}. \modified{Having in mind that we only have positive examples, we have evaluated the results with quality dimensions that are usually applied in process mining: fitness 
and simplicity,} \inserted{which is a common combination (e.g., MML/MDL principles, \citep{wallace1999minimum}) 
in situations with scarcity of data, where other regularisation terms cannot be used.} 

\modified{According to this balance in performance metrics, the results are quite good, showing} that our approach was effective in noise reduction independently of the quality of the examples. However, \modified{the procedure} is very sensitive to the quality in the training set: increasing the quality of the examples during this phase could be translated into a significant reduction of the training set size. Thereby, if the examples are rich enough, the algorithm is able to learn good quality models with fewer examples.  

 In order to explore and  analyse the \ShortNameMiningAlgorithm{}  method to other tasks provided by JIGSAWS dataset, we have developed a Shiny application\footnote{https://safe-tools.dsic.upv.es/shiny/SurgicalWorkflowMining/}. \modified{On this web application, users} can select a task, apply the learning models and replay the different examples (trials) on the obtained models. It also possible to watch all the steps performed by the 
 algorithm until obtaining the final  models.

\inserted{
In the future we will concentrate on extending our approach 
to also consider short descriptions in natural language of the task (performed by an expert) in order to provide the system with an automatic or semiautomatic verification step after the learning process. Finally, we will apply our developments to the automatic learning and supervision system of skilled tasks that is one of the most relevant application areas of our method.
}


\section*{Acknowledgements}
This work has been partially supported by the EU (FEDER) and the Spanish MINECO under grants TIN2014-61716-EXP (SUPERVASION) and RTI2018-094403-B-C32, and by Generalitat Valenciana under grant PROMETEO/2019/098. David Nieves is also supported by the Spanish MINECO under FPI grant (BES-2016-078863).









\section*{References}




\bibliography{biblio}

\end{document}